\documentclass[11pt,letterpaper]{article}
\usepackage[left=1in,right=1in,top=1in,bottom=1in]{geometry}

\usepackage[utf8]{inputenc} 
\usepackage[T1]{fontenc}    
\usepackage{hyperref}       
\usepackage{url}            
\usepackage{booktabs}       
\usepackage{nicefrac}       
\usepackage{microtype}      
\usepackage{xcolor}         

\usepackage{caption} 

\usepackage{setspace}
\usepackage{amsmath, amssymb, amsfonts, enumerate, verbatim, bbm, graphicx}
\usepackage{algorithm}
\usepackage{subfig}
\usepackage[noend]{algpseudocode}
\usepackage[parfill]{parskip}

\makeatletter
\def\BState{\State\hskip-\ALG@thistlm}
\makeatother

\newtheorem{lemma}{Lemma}
\newtheorem{fact}{Fact}
\newtheorem{theorem}{Theorem}
\newtheorem{proposition}{Proposition}

\newtheorem{model}{Model}

\title{Bandits with Dynamic Arm-acquisition Costs\footnote{Appeared in the \emph{{58}\textsuperscript{th} Annual Allerton Conference on Communication, Control, and Computing (Allerton 2022).}}}

\author{%
  Anand Kalvit\textsuperscript{1} and Assaf Zeevi\textsuperscript{2} \\
  Columbia University \\
  \texttt{\{\textsuperscript{1}akalvit22,\textsuperscript{2}assaf\}@gsb.columbia.edu}
}

\date{}

\begin{document}

\maketitle

\begin{abstract}
We consider a bandit problem where at any time, the decision maker can add new arms to her consideration set. A new arm is queried at a cost from an \emph{arm-reservoir} containing finitely many \emph{arm-types}, each characterized by a distinct mean reward. The cost of query reflects in a diminishing probability of the returned arm being optimal, unbeknown to the decision maker; this feature encapsulates defining characteristics of a broad class of operations-inspired online learning problems, e.g., those arising in markets with churn, or those involving allocations subject to costly resource acquisition. The decision maker's goal is to maximize her cumulative expected payoffs over a sequence of $n$ pulls, oblivious to the statistical properties as well as types of the queried arms. We study two natural modes of endogeneity in the reservoir distribution, and characterize a necessary condition for achievability of sub-linear regret in the problem. We also discuss a UCB-inspired adaptive algorithm that is long-run-average optimal whenever said condition is satisfied, thereby establishing its tightness. 
\end{abstract}

\section{Introduction}

\textbf{Background.} The stochastic multi-armed bandit (MAB) problem \cite{lai} is a classical machine learning paradigm that provides a succinct abstraction of the \emph{exploration-exploitation} trade-offs inherent to sequential decision making under uncertainty. This problem has guided theory and practice over the years, and appears as a basal motif in several complex data-driven applications; see \cite{bubeck2012regret} for a survey of well-studied formulations. In its simplest formulation, the decision maker must pull at each time instant $t\in\{1,2,...\}$ one out of a set of $K$ possible arms with unknown reward distributions (and any statistical property thereof) in a way that maximizes her cumulative expected payoffs over a sequence of $n$ pulls. A natural metric to assess the performance of an algorithm in this setting is its \emph{regret} w.r.t. an oracle that recommends an optimal arm at each step. The seminal work \cite{lai} was the first to show that the best achievable regret in ``well-separated'' instances of the problem is $\Theta\left( \log n \right)$.

\textbf{Towards countably many arms.} In several applications of MAB, the complexity of decision making resides not only in identifying an optimal arm among $K$ competing alternatives, but is also attributable to the choice of allocating effort across said alternatives and certain ``outside options;'' this characteristic is usually referred to as \emph{choice overload} in decision theory. One natural approach in the literature for modeling choice overload in MAB involves endowing the problem with an \emph{arm-reservoir} that returns upon query ``new'' arms sampled from a distribution over \emph{arm-types}, see, e.g., \cite{berry1997bandit}. Among problems of this nature, a simple yet instructional setting is one where the reservoir only holds finitely many arm-types \cite{kalvit2020finite}. Such finite-typed models appear frequently also in learning problems in the broader operations research literature, see, e.g., \cite{johari2021matching}.

\textbf{Cost of reservoir access.} In MAB settings where choice overload is a defining characteristic, the decision maker may experience elevated costs of acquiring new arms as time progresses. This can be viewed in light of plausible deterioration in the ``quality'' of the reservoir either with increasing number of queries or with time. The former can be interpreted as a Lagrangian relaxation to an optimization problem with capacity constraints on reservoir queries; a setting potentially of interest to online resource allocation problems under costly resource acquisition. The latter aspect has connotations related to market churn. For example, the relative abundance of agents of different types in a market may undergo temporal variations over the platform's planning horizon. This is the case, for example, in online service-platforms, where arms may represent agents capable of abandoning the platform if kept idle for protracted durations. These effects may catalyze an agent-departure process and a temporal non-stationarity that may seriously hinder the decision maker's ability to discern ``good'' arms from ``inferior'' ones. There has been significant recent interest in non-stationary bandit models, however, most of the literature is limited largely to non-stationarities in rewards, and antecedents on non-stationarities in arm-reservoirs are markedly absent (see the literature review in \S\ref{sec:literature}). In this paper, we provide the first systematic treatment of this aspect by investigating the statistical limits of learning under endogenous variations in the arm-reservoir distribution.

\textbf{Stylized model.} To distill key insights, it will be convenient to focus on the paradigmatic case of exactly two arm-types in the reservoir; this serves to highlight statistical idiosyncrasies of the problem without unnecessary mathematical detail. The complexity of the \emph{static} version of the problem is driven by three primitives: (i) the gap $\Delta$ between the mean rewards associated with the two arm-types; (ii) the fraction $\alpha$ of the ``optimal'' arm-type in the reservoir; and (iii) the horizon of play $n$. The feature of costly arm-acquisitions is incorporated by endogenizing $\alpha$; consequently, its evolution is given by the stochastic process $\left( \alpha(t) : t =1,2,... \right)$. Thus, at any time $t$, the probability that a query of the arm-reservoir returns an arm of the optimal type is given by (the sample path-dependent random variable) $\alpha(t)$. As we shall later see, the interesting regime is where $\alpha(t)$ becomes vanishingly small as $t$ grows; this captures departure of optimal (unexplored) arms from the reservoir, and abstracts out key characteristics of many online platforms that serve a large population of impatient agents.

\textbf{Challenges due to costly reservoir access.} It is non-trivial to design ``good'' policies that are reservoir distribution-agnostic as well as gap-adaptive. To see this, consider the simplest scenario where $\alpha(t)=c$ (constant), and the decision maker is endowed with ex ante knowledge of $c$ as well as the horizon $n$. A natural heuristic in this setting is to query $\Omega\left( c^{-1}\log n \right)$ arms upfront (this would guarantee with probability $\Omega\left( 1/n \right)$ the existence of at least one optimal arm among the queried ones), and subsequently deploy a conventional bandit algorithm such as Thompson Sampling \cite{agrawal2012analysis} or UCB \cite{auer2002} on the collected set of arms. One can quite easily show that such an approach will incur poly-logarithmic regret in $n$. On the other hand, it is possible to achieve $\tilde{\mathcal{O}}\left( \left(c\Delta\right)^{-1} \log n\right)$ regret (up to poly-logarithmic factors in $\Delta^{-1}$) using a more sophisticated policy \cite{de2021bandits}. Regardless, none of these approaches utilize the fact that there exist exactly $K$ arm-types in the reservoir. By leveraging this knowledge, $c$-dependence of the aforementioned upper bound can, in fact, be relegated to sub-logarithmic terms, surprisingly, in a manner that is adaptive to $c$ \cite{kalvit2020finite}. It is noteworthy, however, that the performance of such approaches is fragile w.r.t. the premise that $\alpha(t)$ remains bounded away from $0$ by some problem-independent constant $c$ (which may or may not be known a priori) at all times. Naturally, endogenous variations potentially causing $\alpha(t)$ to vanish in $t$ will only exacerbate the problem, and it remains unclear if it is even possible in this setting to achieve sub-linear regret relative to the classical full-information benchmark that prescribes pulling an optimal arm at each $t$.

{\bf Contributions.} We first derive a necessary condition for ``complete learning'' when the evolution of $\alpha(t)$ is independent of the decision maker's actions; specifically, $\sum_{t=1}^{\infty}\alpha(t)=\infty$ is \emph{necessary} for achieving sub-linear regret relative to an oracle that knows the identities of optimal arms ex ante (Theorem~\ref{thm:rate}). We also establish its near-tightness in that a slightly stronger version of said condition is sufficient for a gap-\emph{aware} policy based on the Explore-then-Commit principle to achieve poly-logarithmic instance-dependent regret in the problem (Theorem~\ref{thm:alg1-exo}). In addition, we discuss a novel gap-\emph{adaptive} policy based on the UCB principle that achieves a polynomial regret in the same regime (Theorem~\ref{thm:alg2-exo}). We then consider the setting where $\alpha(t)$ is endogenous (policy-dependent), and characterize matching necessary and sufficient conditions (up to leading order terms) for asymptotic-optimality of aforementioned policies. 

Before proceeding with a formal description of our model and discussion of results, we provide a brief overview of related literature below.


\subsection{Related literature}
\label{sec:literature}

{\bf Bandits with state-dependent rewards.} The earliest work on MAB problems involving endogenous arm-reward distributions dates back to the seminal work \cite{gittins1974dynamic} which studied finite-state Markovian bandits. In this model, the state of an arm only changes upon execution of a pull while remaining unchanged otherwise, thus prompting the name \emph{resting bandits}; the celebrated \emph{Gittins index policy} is well-known to maximize the infinite-horizon discounted cumulative expected reward in this setting. In contrast, in the so-called \emph{restless bandits} formulation \cite{whittle1988restless}, the states of all the arms may change simultaneously irrespective of which arms are pulled; in addition, this formulation permits pulling any fixed number of arms in each period.
Subsequent works such as \cite{weber1990index} focus on heuristics that are optimal in an asymptotic regime where the number of arms pulled in each period scales linearly with the total number of arms. More recently, a finite-horizon variant of the restless bandits problem was studied in \cite{frazier2017} under a similar scaling; see \cite{gittins2011multi,zayas2019asymptotically,brown2020index}, etc., for a survey of other well-studied variations.

Our work is quite distinct from this strand of literature: (i) asymptotic analyses in aforementioned papers are w.r.t. the number of arms, not the horizon of play; (ii) the number of arms pulled in each period is fixed at $1$ in our problem setting and does not scale with the total number of arms; (iii) most importantly, cited references formulate the problem as a Markov decision process assuming full knowledge of the transition kernels. In contrast, we consider a learning theoretic formulation where the decision maker is oblivious to the statistical properties of reward distributions as well as the nature of endogeneity in the arm-reservoir.

{\bf Bandits with non-stationary rewards.} This line of work focuses on policies that minimize the expected cumulative regret relative to a \emph{dynamic oracle} that plays at each time $t$ an arm with the highest mean reward at $t$. Some of the early work in this area is premised on a formulation in which the identity of the best arm may change a finite number of times adversarially during the horizon of play, see, e.g., \cite{auer2002nonstochastic}. While other works such as \cite{slivkins2008adapting} study specific models of temporal variation where, for example, rewards evolve according to a Brownian motion, much of the traditional literature is limited, by and large, to a finite number of changes in the mean rewards; see \cite{garivier2011upper} and references therein. Subsequently, a unified framework for studying aforementioned problem classes was provided in \cite{besbes2014stochastic} by introducing a \emph{variation budget} to bound the evolution of mean rewards over the horizon of play. Several other forms of non-stationarity have also been studied in the literature; these include formulations with rotting \cite{levine2017rotting}, recharging \cite{kleinberg2018recharging}, and delay-dependent rewards \cite{cella2020stochastic}, among others (see \cite{besbes2019optimal} for a survey). 

Aforementioned works are largely limited to the study of finite-armed bandit problems where non-stationarity can be ascribed to changes in \emph{arm-means}. In contrast, our work differs fundamentally in that it is premised on an infinite-armed formulation with non-stationarity attributable to an endogenous \emph{arm-reservoir}. In a nutshell, while preceding work focuses on distributional shifts in rewards given a fixed set of arms, we propose a new paradigm where the arm-reservoir itself undergoes distributional shifts owing to possible ``leakages,'' which is functionally a very different concept.

{\bf Bandits with infinitely many arms.} These problems involve settings where an unlimited supply of arms is governed by some \emph{fixed} distribution over an \emph{uncountable} set of arm-types (possible mean rewards); a common reward statistic (usually the mean) uniquely characterizes each arm-type. The infinite-armed bandit problem traces its roots to \cite{berry1997bandit} where it was studied under the Bernoulli reward setting with means distributed Uniformly on $[0,1]$. More general reward and reservoir distributions on $[0,1]$ have also been studied in subsequent works, see, e.g., \cite{wang2009,bonald2013two,carpentier2015simple,chan2018infinite}. Our model differs from this line of work in that we only assume the reservoir to be \emph{finitely supported} and posit no distributional knowledge thereof, unlike cited references. Furthermore, the distribution of arm-types is allowed to vary endogenously over the problem horizon in a manner that may be unknown ex ante.

\subsection{Outline of the paper}

A formal description of the model is provided in \S\ref{sec:formulation}. \S\ref{sec:algorithms} discusses reservoir distribution-agnostic algorithms; two natural modes of endogeneity in the reservoir distribution are discussed in \S\ref{sec:reservoir-models} along with corresponding guarantees on achievable performance. \S\ref{sec:remarks} provides concluding remarks. Proofs and ancillary results are delegated to the technical appendix.

\section{Problem formulation}
\label{sec:formulation}

\textbf{Primitives.} There are finitely many possible arm-types in the reservoir denoted by the collection $\mathcal{K}$; the vector $\boldsymbol{\mu} := \left( \mu_{k} : k\in\mathcal{K} \right)$ characterizes the mean reward (pairwise distinct) associated with each. The decision maker (DM) only knows the cardinality of $\mathcal{K}$; for simplicity of exposition, we assume $\left\lvert \mathcal{K} \right\rvert = 2$ in this work and index the two arm-types by ``1'' and ``2,'' i.e., $\mathcal{K}=\{1,2\}$. Without loss of generality, arm-type~$1$ is assumed ``optimal'' with a gap (or separation) of $\Delta:= \mu_1-\mu_2 > 0$ from the ``inferior'' type (arm-type~$2$); as we shall later see, $\Delta$ is an important driver of the problem's statistical complexity. DM must play one arm at each time $t\in\{1,...,n\}$, where $n$ denotes the horizon of play.

\textbf{Reservoir access.} The collection of arms to have been played at least once until time $t$ (inclusive) is denoted by $\mathcal{I}_t$ (with $\mathcal{I}_0 := \phi$). The set of available actions at $t$ is given by $\mathcal{A}_t= \mathcal{I}_{t-1}\cup\{\text{new}_t\}$; DM must either play an arm from $\mathcal{I}_{t-1}$ or select the action ``$\text{new}_t$'' which corresponds to playing a \emph{new} arm queried from the reservoir. A newly queried arm at time $t$ is optimal-typed with probability $\alpha(t)$ and inferior-typed otherwise ($\alpha(t)$ may potentially be random and endogenous, unbeknown to DM; a detailed discussion is deferred to \S\ref{sec:reservoir-models}). Arm-types are private attributes and remain unobservable throughout. Upon pulling an arm labeled $i$ (henceforth referred to as arm~$i$) for the $j^{\text{th}}$ time, DM observes a $[0,1]$-valued stochastic reward denoted by $X_{i,j}$. The realized rewards are independent across arms and time, and mean-preserving in time (not necessarily identically distributed) keeping the arm fixed. 

\textbf{Admissible controls.} A policy $\pi := \left(\pi_1, \pi_2, ...\right)$ is an adaptive allocation rule that prescribes at time $t$ an action $\pi_t$ (possibly randomized) from $\mathcal{A}_t$. The collection of all observable information until $t$ is given by the \emph{natural filtration} $\mathcal{F}_t := \sigma\left\lbrace \left( \pi_s \right)_{1\leqslant s \leqslant t}, \left\lbrace \left( X_{i,j} \right)_{ 1 \leqslant j \leqslant N_i(t)} : i\in\mathcal{I}_t \right\rbrace \right\rbrace$ (with $\mathcal{F}_0:=\phi$), where $N_i(t)$ indicates the number of times arm~$i$ is pulled until $t$. The cumulative \emph{regret} of $\pi$ after $n$ plays is given by $R_n^\pi := \sum_{t=1}^{n}\left(\mu_1 - X_{\pi_t, N_{\pi_t}(t)} \right)$ and its cumulative \emph{pseudo-regret} by $\mathcal{R}_n^\pi := \sum_{t=1}^{n}\left(\mu_1 - \mu_{\kappa\left(\pi_t\right)}\right)$, where $\kappa\left(\pi_t\right)\in\mathcal{K}$ encodes the type of arm~$\pi_t$; note that both regret as well as pseudo-regret are sample path-dependent by definition. DM is interested in the following stochastic minimization problem
\begin{align}
\inf_{\pi\in\Pi}\mathbb{E}R_n^\pi &= \inf_{\pi\in\Pi}\mathbb{E}\left[ \sum_{t=1}^{n}\left(\mu_1 - X_{\pi_t, N_{\pi_t}(t)} \right) \right] \underset{\mathrm{(\dag)}}{=} \inf_{\pi\in\Pi}\mathbb{E}\left[ \sum_{t=1}^{n}\left(\mu_1 - \mu_{\kappa\left(\pi_t\right)}\right) \right] = \inf_{\pi\in\Pi}\mathbb{E}\mathcal{R}_n^\pi, \label{eqn:problem_formulation}
\end{align}
where the infimum is over the class $\Pi$ of \emph{non-anticipating} policies, i.e., $\pi_t : \mathcal{F}_{t-1}\to \mathcal{P}\left( \mathcal{A}_{t} \right);\ t\in\{1,2,...\}$ ($\mathcal{P}\left( \mathcal{A}_{t} \right)$ denotes the probability simplex on $\mathcal{A}_t$), the expectations are w.r.t. all the possible sources of stochasticity in the problem (rewards, policy, and the reservoir distribution), and $(\dag)$ holds since cumulative regret and pseudo-regret are equal in expectation in the \emph{stochastic bandits} setting (follows from the Tower property of expectations). 

\section{Distribution-agnostic policies}
\label{sec:algorithms}

As discussed in the introduction, our goal in this work is to investigate achievable regret under algorithms that are agnostic to the reservoir distribution. To this end, we discuss two such algorithms; one based on the forced-exploration principle, and the other on optimism-under-uncertainty.

\subsection{A fixed-design policy based on forced exploration}

The non-adaptive ETC (Explore-then-Commit) approach outlined below is predicated on ex ante knowledge of the problem horizon $n$ (this is not a constraining factor since the exponential doubling trick \cite{doubling} can be used to make the algorithm horizon-free) and a gap parameter $\delta\in\left(0,\Delta\right]$. 
In what follows, a \emph{new} arm refers to one that is freshly queried from the arm-reservoir. 

\begin{algorithm}[H]
\caption{\texttt{ALG1}$\left(n,\delta\right)$ \texttt{(Non-adaptive ETC)}}
\label{alg:ALG1}
\begin{algorithmic}[1]
\BState \textbf{Input:} Horizon of play $n$, gap parameter $\delta$. 
\State Set $m = \left\lceil 2\delta^{-2}\log n \right\rceil$.
\BState \textbf{New epoch:} Play $K=2$ \emph{new} arms from the reservoir (call consideration set $\mathcal{A}=\{1,2\}$).
\State Observe rewards $\left( X_{1,1}, X_{2,1} \right)$.
\State Play each arm $m-1$ times more; observe rewards $\left( X_{1,j}, X_{2,j} : j = 2,...,m \right)$.
\If {$\left\lvert {\sum_{j=1}^{m}(X_{1,j}-X_{2,j})} \right\rvert < \delta m$}
\State Permanently discard $\mathcal{A}$ and repeat from step $(3)$.
\Else
\State Permanently commit to arm $i^{\ast} \in \arg\max_{i\in\mathcal{A}}\left\lbrace \sum_{j=1}^{m}X_{i,j} \right\rbrace$.
\EndIf
\end{algorithmic}
\end{algorithm}

\textbf{Policy dynamics.} The horizon is divided into epochs of length $2m=\Theta\left( \log n \right)$ each. In each epoch, the algorithm re-initializes by querying the arm-reservoir for a pair of new arms, and playing them $m$ times each. Subsequently, the pair is classified as either ``distinct'' or ``identical''-typed via a hypothesis test (step $6$ of Algorithm~\ref{alg:ALG1}). If classified as distinct, the algorithm commits the residual budget of play to the empirically superior arm among the two (with ties broken arbitrarily). On the other hand, if the pair is classified as identical, the algorithm discards it permanently and ushers in a new epoch. The entire process repeats until a distinct-typed pair is identified. \texttt{ALG1}$(n,\delta)$, albeit non-adaptive to $\Delta$, serves as an insightful basal motif for algorithm design and its operating principle will guide the development of the $\Delta$-adaptive algorithm discussed next.

\subsection{A UCB-based approach with adaptive resampling}

In this section, we revisit the UCB-based adaptive policy proposed in \cite{kalvit2020finite} for the \emph{static} version of our problem where $\alpha(t)$ is constant. This policy is restated as \texttt{ALG2} after suitable modifications for reasons discussed next. The original policy (Algorithm~2 in cited reference) suffers a limitation through its dependence on ex ante knowledge of the support of reward distributions. In particular, the algorithm requires reward distributions associated with the arm-types to be ``maximally supported'' on $[0,1]$, e.g., only distributions such as Bernoulli$(\cdot)$, Uniform on $[0,1]$, Beta$(\cdot,\cdot)$, etc., are amenable to its performance guarantees; Uniform on $[0,0.5]$, on the other hand, is not. We identify a simple fix to this issue and propose adding a centered Gaussian noise term to the cumulative-difference-of-reward statistic (see step~(5) of Algorithm~\ref{alg:ALG2}) to essentially create an unbounded support. This rids the algorithm of its dependence on assumptions pertaining to the support, and at the same time also preserves its regret guarantees in the static setting. In fact, our modifications lead to a sharper characterization of the scaling behavior of regret w.r.t. $\Delta$ in the static setting (see Theorem~\ref{thm:alg2-exo} and \ref{thm:alg2-endo}). 

\textbf{Remark.} The original policy (Algorithm~2 in cited reference) uses a threshold that is distinct from $4\sqrt{m\log m}$ (see step~(5) of Algorithm~\ref{alg:ALG2}); the choice of $4\sqrt{m\log m}$ here aims to simplify the exposition of key technical innovations and facilitate an easy comparison between various upper bounds.

\begin{algorithm}[H]
\caption{\texttt{ALG2 (Nested UCB)}}
\label{alg:ALG2}
\begin{algorithmic}[1]
\BState \textbf{New epoch ($\boldsymbol{t\gets 0}$):} Play $K=2$ \emph{new} arms from the reservoir (call set $\mathcal{A}=\{1,2\}$).
\State Observe rewards $\left( X_{1,1}, X_{2,1} \right)$.
\State Minimum per-arm sample count $m\gets 1$.
\State Generate an independent sample of a standard Gaussian distribution $\mathcal{Z}$.
\For{$t \in\{3,4,...\}$}
\If {$\left\lvert \mathcal{Z} + \sum_{j=1}^{m}\left( X_{1,j}-X_{2,j} \right) \right\rvert < 4\sqrt{m\log m}$}
\State Permanently discard $\mathcal{A}$ and repeat from step (1).
\Else
\State Play arm~$i_t\in\arg\max_{i\in\mathcal{A}}\left( \frac{\sum_{j=1}^{N_i(t-1)}X_{i,j}}{N_i(t-1)} + \sqrt{\frac{2\log(t-1)}{N_i(t-1)}} \right)$.
\State Observe reward $X_{i_t,N_{i_t}(t)}$.
\If {$m < \min_{i\in\mathcal{A}} N_i(t)$}
\State $m \gets m+1$.
\EndIf
\EndIf
\EndFor
\end{algorithmic}
\end{algorithm}

\textbf{Policy dynamics.} \texttt{ALG2} also has an episodic dynamic with exactly one pair of arms played per episode. It is noteworthy that \texttt{ALG2} plays arms according to \texttt{UCB1} \cite{auer2002} in every episode as opposed to playing them equally often until committing to the empirically superior one \`a la \texttt{ALG1}$(n,\delta)$. Secondly, \texttt{ALG2} never ``commits'' to an arm (or a consideration set); the implication is that the algorithm will keep querying new consideration sets from the reservoir throughout the horizon of play. Aforementioned \emph{adaptive resampling} property is at the core of its horizon-free nature. 

\textbf{Operating principle.} At any time, \texttt{ALG2} computes a threshold of $\mathcal{O}\left( \sqrt{m\log m} \right)$ for the length-$m$ cumulative-difference-of-reward process, where $m$ denotes the minimum sample count among the two arms. If the envelope of said process is dominated by $\mathcal{O}\left( \sqrt{m\log m} \right)$, the arms are likely to belong to the same type (simultaneously optimal or inferior). The explanation stems from the Law of the Iterated Logarithm (see \cite{durrett2019probability}, Theorem~8.5.2): a zero-drift length-$m$ random walk process has its envelope bounded by $\mathcal{O}\left( \sqrt{m\log\log m} \right)$. In the aforementioned scenario, the algorithm discards the consideration set and subsequently, a new epoch is ushered in. This is done to avoid the possibility of incurring linear regret in the event that the two arms are inferior, since it is statistically impossible to distinguish a simultaneous-inferior consideration set from one where both arms are optimal, in the absence of any auxiliary information such as the mean rewards associated with the two types. On the other hand, if the cumulative-difference-of-reward dominates $\mathcal{O}\left( \sqrt{m\log m} \right)$, the consideration set is likely to contain arms of distinct types and \texttt{ALG2} continues to run \texttt{UCB1} on this set until the $\mathcal{O}\left( \sqrt{m\log m} \right)$ threshold is breached again.

\textbf{Reason for introducing the Gaussian corruption.} Centered Gaussian noise is added to the cumulative-difference-of-reward process in step (6) of \texttt{ALG2} to avoid the possibility of incurring linear regret should the support of the reward distributions be a ``very small'' subset of $[0,1]$. To illustrate this point, suppose that the rewards associated with the types are deterministic with $\Delta < 2\sqrt{2\log 2}$. Then, as soon as the algorithm queries a heterogeneous consideration set (one arm optimal and the other inferior) and the per-arm sample count reaches $2$, the cumulative-difference-of-reward statistic will satisfy $\left\lvert \sum_{j=1}^{2} \left( X_{1,j} - X_{2,j} \right) \right\rvert = 2\Delta < 4\sqrt{2\log 2}$, resulting in the consideration set getting discarded. On the other hand, if the consideration set is homogeneous (both arms optimal or inferior), the algorithm will still re-initialize within a finite number of samples in expectation (again, owing to the Law of the Iterated Logarithm). This will force the algorithm to keep querying new arms from the reservoir at rate that is linear in time, which is tantamount to incurring linear regret in the horizon. The addition of centered Gaussian noise hedges against this risk by guaranteeing that the cumulative-difference-of-reward process essentially has an infinite support at all times (even when the reward distributions might be degenerate). This rids the regret performance of its fragility w.r.t. ex ante knowledge of the support of reward distributions. The next proposition crystallizes this discussion.

\begin{proposition}[Persistence of heterogeneous consideration sets]
\label{prop}
Let $\left\lbrace X_{i,j} : j=1,2,... \right\rbrace$ be a collection of independent samples from an arm of type~$i$ (implying $\mathbb{E}\left[ X_{1,j} - X_{2,j} \right]=\Delta$ $\forall\ j=1,2,..$). Let $\mathcal{Z}$ be an independently generated standard Gaussian random variable. Then,
\begin{align}
\mathbb{P}\left( \bigcap_{m \geqslant 1} \left\lbrace \left\lvert \mathcal{Z} + \sum_{j=1}^{m} \left( X_{1,j} - X_{2,j} \right) \right\rvert \geqslant 4\sqrt{m\log m} \right\rbrace \right) > \frac{\bar{\Phi}\left( f\left( T_0 \right) \right)}{2} =: \beta_\Delta > 0, \label{eqn:beta} 
\end{align}
where $\bar{\Phi}(v) := 1/\sqrt{2\pi}\int_{v}^{\infty}e^{-u^2/2}du\ \forall\ v>0$ is the right-tail of the standard Gaussian CDF, $T_0 := \left\lceil \left( 64/\Delta^2 \right) \log^2\left( {64}/{\Delta^2} \right) \right\rceil$, and $f(x) := x + 4\sqrt{x\log x}\ \forall\ x \geqslant 1$.
\end{proposition}

\textbf{Interpretation of $\boldsymbol{\beta_\Delta}$.} First of all, note that $\beta_\Delta$ admits a closed-form characterization in terms of standard functions and satisfies $\beta_\Delta > 0$ when $\Delta > 0$ with $\lim_{\Delta\to 0}\beta_\Delta = 0$. Secondly, \emph{$\beta_\Delta$ depends exclusively on $\Delta$}, and represents a lower bound on the probability that \texttt{ALG2} will never discard a consideration set containing arms of distinct types. This meta-result will be key to the upper bounds stated in forthcoming sections. 

\section{Natural models of the arm-reservoir}
\label{sec:reservoir-models}

The probability $\alpha(t)$ of a newly queried arm at time $t$ being optimal-typed will likely vary over the horizon of play in realistic settings. For example, in the context of crowdsourcing applications, the availability of ``high quality'' workers for a given task may depend on the prevailing population-level perception of the platform. This could plausibly be a function of the \emph{age} $t$ of the platform. Alternatively, the reservoir may react to a query at time $t$ through its \emph{cumulative query count} $\mathcal{J}_t$; this model may be suited to settings where, for example, acquiring a new resource is costly and yields diminishing returns. We capture these aspects through two reservoir models described next.



\begin{model}[Exogenous reservoir]
\label{model:exo}
$\left( \alpha(t) : t=1,2,...\right)$ is a non-increasing deterministic process with $\alpha(1) = c\in\left(0,1\right)$, evolving independently of the decision maker's actions.
\end{model}

\begin{model}[Endogenous reservoir]
\label{model:endo}
$\left( \alpha(t) : t=1,2,...\right)$ evolves as $\alpha(t) = g\left( \mathcal{J}_{t-1} \right)$, where $g:\mathbb{N}\cup\{0\}\mapsto(0,c]$ is a non-increasing deterministic mapping with $g(0)=c\in\left(0,1\right)$, and $\mathcal{J}_t$ denotes the number of reservoir queries until time $t$ (inclusive) with $\mathcal{J}_0 := 0$. 
\end{model}


Our first result below states a necessary condition for achievability of sub-linear regret in the two reservoir models.

\begin{theorem}[Necessary conditions for ``complete learning'' in the two reservoir models]
\label{thm:rate}

\begin{enumerate}

\item Under Model~\ref{model:exo}, if $\sum_{t=1}^{\infty}\alpha(t) < \infty$, the expected cumulative regret of any policy $\pi$ grows linearly in the horizon of play, i.e., $\mathbb{E}R_n^{\pi}=\Omega\left( \Delta n \right)$, where the $\Omega(\cdot)$ only hides multiplicative constants independent of $\Delta$ and $n$. Equivalently, a necessary condition for achieving sub-linear regret in the problem is $\sum_{t=1}^{\infty}\alpha(t) = \infty$.

\item Under Model~\ref{model:endo}, if $\sum_{t=1}^{\infty}g(t) < \infty$, the expected cumulative regret of any policy $\pi$ grows linearly in the horizon of play, i.e., $\mathbb{E}R_n^{\pi}=\Omega\left( \Delta n \right)$, where the $\Omega(\cdot)$ only hides multiplicative constants independent of $\Delta$ and $n$. Equivalently, a necessary condition for achieving sub-linear regret in the problem is $\sum_{t=1}^{\infty}g(t) = \infty$.

\end{enumerate}

\end{theorem}

The proof relies essentially on reduction to a full-information setting where the decision maker observes the true mean of an arm immediately upon play (see the technical appendix for details). The optimal policy is this setting will keep querying the reservoir for new arms until it finds one with the optimal mean, which it will subsequently commit the rest of its sampling budget to. The conditions stated in Theorem~\ref{thm:rate} are necessary for this policy to find an optimal arm within its lifetime. It is only natural that the same condition is necessary for achievability of sub-linear regret in the general setting where the decision maker only observes a noisy version of the true means. Surprisingly, however, these conditions are also \emph{almost-sufficient} for sub-linear regret (as forthcoming results will show), and are therefore \emph{nearly-tight}. 

\subsection{Exogenous reservoirs}
\label{sec:exo}


The focus here will be on settings specified by Model~\ref{model:exo}. Theorem~\ref{thm:rate} establishes a necessary condition of $\sum_{t=1}^{\infty}\alpha(t) = \infty$ for achievability of sub-linear instance-dependent regret in the problem. In what follows, we show a slightly more refined characteristic: $\tilde{\Theta}\left(t^{-1}\right)$ is, in fact, a critical rate for ``complete learning'' in that policies achieving sub-linear regret exist if $\alpha(t) = \omega \left( \left(\log t\right)/t\right)$.

To elucidate the criticality of the $\tilde{\Theta}\left(t^{-1}\right)$ rate, it will be convenient to consider a parameterization of $\alpha(t)$ given by $\alpha(t) = ct^{-\gamma}$, where $c$ is as specified in Model~\ref{model:exo} and $\gamma\in[0,1)$ is a fixed parameter. This parameterization offers meaningful insights as to the statistical complexity of the problem w.r.t. $\gamma$ and facilitates an easy comparison between the regret guarantees of the algorithms discussed in \S\ref{sec:algorithms}. We will begin with an upper bound on the performance of Algorithm~\ref{alg:ALG1}. 

\begin{theorem}[Upper bound for \texttt{ALG1}$(n,\delta)$]
\label{thm:alg1-exo}
Under Model~\ref{model:exo} with $\alpha(t) = ct^{-\gamma}$, where $\gamma\in[0,1)$, the expected cumulative regret of the policy $\pi$ given by Algorithm~\ref{alg:ALG1} satisfies
\begin{align*}
\limsup_{n\to\infty}\frac{\mathbb{E}R_n^{\pi}}{\left(\log n\right)^{\frac{1}{1-\gamma}}} \leqslant 24\Delta\left( \frac{8}{\delta^2c} \right)^{\frac{1}{1-\gamma}} \mathfrak{F}\left(\left\lceil \frac{\gamma}{1-\gamma} \right\rceil\right),
\end{align*}
where $\mathfrak{F}(\cdot)$ denotes the \emph{Factorial} function.
\end{theorem}

The proof of Theorem~\ref{thm:alg1-exo} involves technical details beyond the scope of a succinct discussion here (see the technical appendix for details). The above result establishes that the inflation in regret as a result of asymptotically vanishing $\alpha(t)$ is at most poly-logarithmic in the horizon for ``slowly decaying'' $\alpha(t)$, until a sharp phase transition to linear regret occurs around the $\alpha(t) = \tilde{\Theta}\left( t^{-1} \right)$ ``critical rate'' (see Theorem~\ref{thm:rate}.1).

\textbf{Remarks.} \textbf{(i) Logarithmic regret.} Theorem~\ref{thm:alg1-exo} implies that $\mathbb{E}R_n^\pi = \mathcal{O}\left( \left(c\Delta\right)^{-1}\log n \right)$ when $\delta=\Delta$ and the reservoir has no ``leakage'' ($\gamma=0$), consistent with known results for the static version of the problem \cite{kalvit2020finite}. \textbf{(ii) Improving sample-efficiency.}  Instead of discarding both arms after step $6$ of Algorithm~\ref{alg:ALG1}, one can, in principle, discard only one arm, and query only one new arm from the reservoir as replacement. The regret incurred by this modified policy will differ only in absolute constants. The given design only intends to unify Algorithm~\ref{alg:ALG1} structurally with the other algorithm discussed in \S\ref{sec:algorithms} so as to facilitate an easy comparison between the performance guarantees of the two algorithms. 

\begin{theorem}[Upper bound for \texttt{ALG2}]
\label{thm:alg2-exo}
Under Model~\ref{model:exo}, the expected cumulative regret of the policy $\pi$ given by Algorithm~\ref{alg:ALG2} satisfies
\begin{align*}
\limsup_{n\to\infty}\frac{\alpha(n)\mathbb{E}R_n^{\pi}}{\log n} \leqslant \frac{8}{\Delta\beta_\Delta},
\end{align*}
where $\beta_\Delta$ is as defined in \eqref{eqn:beta}.
\end{theorem}

Proof is provided in the technical appendix.

{\bf Discussion.} It follows directly from the above result that $\alpha(n) = \omega \left( {\log n}/{n} \right)$ is sufficient for \texttt{ALG2} to be first-order optimal. On the other hand, we have already identified $\alpha(n) = \omega \left( {1}/{n} \right)$ as a necessary condition for the existence of a first-order optimal policy (Theorem~\ref{thm:rate}.1). Thus, the characterization of $t^{-1}$ as a critical rate in Theorem~\ref{thm:rate}.1 is sharp up to a logarithmic scaling term. 

\textbf{Remark.} The scaling factor $\beta_\Delta$ in Theorem~\ref{thm:alg2-exo} can, in fact, be shaved off entirely by introducing in \texttt{ALG2} an initial ``burn-in'' phase (sub-linear and coercive in the horizon) during each epoch \`a la \texttt{ALG1}$(n,\delta)$. This will, however, be achieved at the expense of \texttt{ALG2}'s \emph{anytime} property. Horizon-independence can then be restored by the use of a standard exponential doubling trick, see, e.g., \cite{doubling}. The resulting algorithm is not discussed in this paper for brevity.


{\bf Comparison of theoretical performance.} To facilitate a direct comparison between the two upper bounds, it is conducive to consider $\alpha(n)=cn^{-\gamma}$ with $\gamma<1$ in Theorem~\ref{thm:alg2-exo}. Evidently, \texttt{ALG2} pays a heavy price for adaptivity to $\Delta$ which reflects in a polynomial $\tilde{\mathcal{O}}\left( n^{\gamma} \right)$ regret as compared to the poly-logarithmic $\mathcal{O}\left( \left( \log n \right)^{\frac{1}{1-\gamma}} \right)$ regret achievable under \texttt{ALG1}$(n,\delta)$. As to whether a performance gap between $\Delta$-aware and $\Delta$-adaptive algorithms is fundamental in Model~\ref{model:exo} remains an open problem at the moment.

\subsection{Endogenous reservoirs}
\label{sec:endo}
 
The focus here will be on settings specified by Model~\ref{model:endo}. We will begin with an upper bound on the theoretical performance of Algorithm~\ref{alg:ALG1}. As before, it will be conducive to pivot to a parametric family of mappings $g(\cdot)$.

\begin{theorem}[Upper bound for \texttt{ALG1}$(n,\delta)$]
\label{thm:alg1-endo}
Under Model~\ref{model:endo} with $g(u) = c(u+1)^{-\gamma}$ for $u \geqslant 0$, where $\gamma\in[0,1)$ is a fixed parameter, the expected cumulative regret of the policy $\pi$ given by Algorithm~\ref{alg:ALG1} satisfies
\begin{align*}
\limsup_{n\to\infty}\frac{\mathbb{E}R_n^{\pi}}{\log n} \leqslant \left( \frac{48\Delta}{\delta^2} \right) \left( \frac{4}{c} \right)^{\frac{1}{1-\gamma}} \mathfrak{F}\left(\left\lceil \frac{\gamma}{1-\gamma} \right\rceil\right),
\end{align*} 
where $\mathfrak{F}(\cdot)$ denotes the Factorial function.
\end{theorem}

Proof is provided in the technical appendix. Evidently, unlike Theorem~\ref{thm:alg1-exo}, only the multiplicative factors of the logarithmic term in Theorem~\ref{thm:alg1-endo} blow up as $\gamma$ approaches $1$. Thus, under Model~\ref{model:endo}, a sharper phase transition from logarithmic to linear regret occurs at the critical reservoir-depletion rate of $\tilde{\Theta}\left( u^{-1} \right)$, where $u$ is the cumulative query count. We next look at the performance guarantee of \texttt{ALG2} under Model~\ref{model:endo}.

\begin{theorem}[Upper bound for \texttt{ALG2}]
\label{thm:alg2-endo}
Under Model~\ref{model:endo}, the expected cumulative regret of the policy $\pi$ given by Algorithm~\ref{alg:ALG2} satisfies
\begin{align*}
\limsup_{n\to\infty}\frac{\mathbb{E}R_n^{\pi}}{\log n} \leqslant \left( \frac{16c}{\Delta} \right)\sum_{k=0}^{\infty} \exp \left( - \beta_\Delta \sum_{j=0}^{k-1} g\left(2j\right) \right),
\end{align*}
where $\beta_\Delta$ is as defined in \eqref{eqn:beta}.
\end{theorem}

Proof is provided in the technical appendix.

\textbf{Remark.} If the function $g(\cdot)$ is constant (equal to $c$), the upper bound in Theorem~\ref{thm:alg2-endo} is bounded above by $32/\left(\Delta\beta_\Delta\right)$, which is independent of $c$. The implication is that achievable regret in the problem depends on the probability $c$ of sampling optimal-typed arms from the reservoir, surprisingly, only through sub-logarithmic terms when the reservoir is ``static.'' This is fundamentally distinct from the upper bound in Theorem~\ref{thm:alg1-endo} which scales inversely with $c$ when $\gamma=0$.

{\bf Comparison of theoretical performance.} Akin to Theorem~\ref{thm:alg2-exo}, the $\beta_\Delta$ factor in Theorem~\ref{thm:alg2-endo} may also be shaved off, albeit at the expense of \texttt{ALG2}'s anytime property. Then, for $g(u) = c(u+1)^{-\gamma}$, the upper bound in Theorem~\ref{thm:alg2-endo} can be shown to be bounded above by $\left( c/\Delta \right) \left( 4/ c \right)^{1/(1-\gamma)}\mathfrak{F}\left( \left\lceil \gamma/(1-\gamma) \right\rceil \right)$. This order matches (up to numerical factors) the upper bound in Theorem~\ref{thm:alg1-endo}. However, as $\gamma$ approaches $0$, the upper bound in Theorem~\ref{thm:alg1-endo} approaches a scaling of $c^{-1}$ while that in Theorem~\ref{thm:alg2-endo} is independent of $c$ in the limit. This observation suggests that there is merit to using \texttt{ALG2} when the reservoir is \emph{static} or \emph{nearly-static}, i.e., when it suffers a negligible leakage (or loss) of optimal arms over time (or with increasing number of queries).

In a nutshell, the upper bounds in Theorem~\ref{thm:alg1-endo} and \ref{thm:alg2-endo}, combined with the lower bound in Theorem~\ref{thm:rate}.2, underscore the criticality of $g(u) = \tilde{\Theta}\left( u^{-1} \right)$ for achievability of sub-linear regret in the problem when there is attrition of optimal-typed arms from the reservoir with increasing number of queries.

\section{Concluding remarks}
\label{sec:remarks}

This work attempts to develop a principled approach to understanding the impact of endogenous variations in countably many-armed bandit problems through stylized arm-reservoir models. While this paper provides a sharp characterization of critical reservoir-decay rates for achieving sub-linear regret in two natural arm-reservoir models, an important outstanding challenge is to identify more ``reasonable'' models of endogeneity, and the design of ``robust'' algorithms for such settings. To our best knowledge, the modeling and methodological approaches adopted in this paper are novel; we hope these may guide future research on some of the interesting directions emanating out of this work.

\bibliographystyle{acm}
\bibliography{ref}

\newpage

\appendix

\section*{Technical appendix: General organization}

\begin{enumerate}
\item Appendix~\ref{proof:rate} provides the proof of Theorem~\ref{thm:rate}.
\item Appendix~\ref{appendix:prop} provides the proof of Proposition~\ref{prop}.
\item Appendix~\ref{sec:analysis-alg1} provides the regret analysis framework for Algorithm~\ref{alg:ALG1}.
\item Appendix~\ref{proof:alg1-exo} provides the proof of Theorem~\ref{thm:alg1-exo}.
\item Appendix~\ref{proof:alg1-endo} provides the proof of Theorem~\ref{thm:alg1-endo}.
\item Appendix~\ref{sec:aux-alg2} states the auxiliary results used in regret analysis of Algorithm~\ref{alg:ALG2}.
\item Appendix~\ref{sec:analysis-alg2} provides the regret analysis framework for Algorithm~\ref{alg:ALG2}.
\item Appendix~\ref{proof:alg2-exo} provides the proof of Theorem~\ref{thm:alg2-exo}.
\item Appendix~\ref{proof:alg2-endo} provides the proof of Theorem~\ref{thm:alg2-endo}.
\end{enumerate}

\section{Proof of Theorem~\ref{thm:rate}}
\label{proof:rate}

\subsection{Proof for Model~\ref{model:exo}}

In order to prove this result, we consider an oracle that can perfectly observe whether an arm is ``optimal'' or ``inferior''-typed immediately upon pulling it. If such an oracle incurs linear regret, then every policy that only gets to observe a noisy realization of the mean rewards associated with the types, must necessarily incur linear regret as well. 

Clearly, the optimal oracle policy $\pi^{\ast}$ is one that keeps pulling new arms until it finds one of the optimal type (type~1), which it then persists with for the remaining duration of play. Let $Y$ denote the time at which an arm of the optimal type is pulled for the first time under $\pi^{\ast}$. Then,
\begin{align}
\mathbb{P}\left( Y \geqslant k \right) = \prod_{t=1}^{k-1}\left( 1 - \alpha(t) \right)\ \ \ \text{for}\ k\geqslant 2,\ \ \ \  \mathbb{P}\left( Y \geqslant 1 \right) = 1. \label{eqn:prob}
\end{align}

\noindent The expected cumulative regret of the aforementioned policy at time $n$ is
\begin{align*}
\mathbb{E}R_{n}^{\pi^{\ast}} = \sum_{k=1}^{n}\mathbb{P}\left( Y = k \right) \Delta(k-1) + \mathbb{P}\left( Y > n \right)\Delta n > \mathbb{P}\left( Y > n \right)\Delta n > \mathbb{P}\left( Y = \infty \right)\Delta n.
\end{align*}

Thus, if $\mathbb{P}\left( Y = \infty \right)$ is bounded away from $0$, linear regret is unavoidable. Since $\lim_{t\to\infty}\alpha(t) = 0$, we know that $\exists\ t_0\in\mathbb{N}$ s.t. $\alpha(t) < 1/2$ for all $t>t_0$. Then,
\begin{align}
\mathbb{P}\left( Y = \infty \right) = \prod_{t=1}^{\infty}\left( 1 - \alpha(t) \right) = \exp\left( \sum_{t=1}^{\infty}\log \left( 1 - \alpha(t) \right) \right) &= \prod_{t=1}^{t_0}\left( 1 - \alpha(t) \right)\exp\left( \sum_{t=t_0+1}^{\infty}\log \left( 1 - \alpha(t) \right) \right) \notag \\
&> \prod_{t=1}^{t_0}\left( 1 - \alpha(t) \right)\exp\left( -2\sum_{t=t_0+1}^{\infty} \alpha(t) \right), \label{eqn:prob2}
\end{align}
where the final inequality follows since $\alpha(t)<1/2$ for $t>t_0$ and $\log(1-x) > -2x$ for $x\in\left(0,1/2\right]$. Since $t_0$ is finite, it is clear from \eqref{eqn:prob2} that a sufficient condition for $\mathbb{P}\left( Y = \infty \right)$ to be bounded away from $0$ is the summability of $\alpha(t)$, i.e., $\sum_{t\in\mathbb{N}}\alpha(t) < \infty$. \hfill $\square$

\subsection{Proof for Model~\ref{model:endo}}

The proof for this model proceeds along similar lines as the above. One starts by considering an oracle that observes the type of a queried arm perfectly and immediately. The structure of the optimal oracle policy can be argued to be identical to the one discussed in the previous section. Subsequent steps of the proof are instructive. \hfill $\square$

\section{Proof of Proposition~\ref{prop}}
\label{appendix:prop}

Consider the following stopping time:
\begin{align*}
\tau := \inf \left\lbrace m \in\mathbb{N} : \mathcal{Z} + \sum_{j=1}^{m}\left( X_{1,j} - X_{2,j} \right) < 4\sqrt{m\log m} \right\rbrace.
\end{align*}
Since $\mathbb{P}\left( \bigcap_{m \geqslant 1}\left\lbrace \left\lvert \mathcal{Z} + \sum_{j=1}^{m}\left( X_{1,j} - X_{2,j} \right) \right\rvert \geqslant 4\sqrt{m\log m} \right\rbrace \right) \geqslant \mathbb{P}(\tau = \infty)$, it suffices to show that $\mathbb{P}(\tau=\infty)$ is bounded away from $0$. To this end, define the following:
\begin{align*}
T_0 &:= \left\lceil \left( \frac{64}{\Delta^2} \right)\log^2\left( \frac{64}{\Delta^2} \right) \right\rceil, \\
f(x) &:= x + 4\sqrt{x\log x} & \mbox{for $x\geqslant 1$}.
\end{align*}

\begin{lemma}
\label{lemma:prop1}
It is the case that
\begin{align*}
\left\lbrace \mathcal{Z} > f\left( T_0 \right) \right\rbrace \subseteq \bigcap_{m=1}^{T_0}\left\lbrace \mathcal{Z} + \sum_{j=1}^{m}\left( X_{1,j} - X_{2,j} \right) \geqslant 4\sqrt{m\log m} \right\rbrace.
\end{align*}
\end{lemma}

\textbf{Proof of Lemma~\ref{lemma:prop1}.} Note that 
\begin{align*}
\mathcal{Z} &> f\left( T_0 \right) \\
&= T_0 + 4\sqrt{T_0\log T_0} \\
&\geqslant m + 4\sqrt{m\log m}\ \forall\ 1 \leqslant m \leqslant T_0 \\
&\underset{\mathrm{(\mathfrak{a})}}{\geqslant} \sum_{j=1}^{m}\left( X_{2,j} - X_{1,j} \right) + 4\sqrt{m\log m}\ \forall\ 1 \leqslant m \leqslant T_0 \\
\implies \mathcal{Z} + \sum_{j=1}^{m}\left( X_{1,j} - X_{2,j} \right) &\geqslant 4\sqrt{m\log m}\ \forall\ 1 \leqslant m \leqslant T_0,
\end{align*}
where $(\mathfrak{a})$ follows since the rewards are bounded in $[0,1]$, i.e., $\left\lvert X_{1,j} - X_{2,j} \right\rvert \leqslant 1$. \hfill $\square$

\begin{lemma} 
\label{lemma:prop2}
For $m \geqslant T_0$, it is the case that
\begin{align*}
\Delta \geqslant 8\sqrt{\frac{\log m}{m}}.
\end{align*}
\end{lemma}

\textbf{Proof of Lemma~\ref{lemma:prop2}.} First of all, note that $T_0 \geqslant 64$ (since $\Delta \leqslant 1$). For $s = \left( {64}/{\Delta^2} \right)\log^2\left( {64}/{\Delta^2} \right)$, one has 
\begin{align*}
\Delta^2 = \frac{64\log^2\left( \frac{64}{\Delta^2}\right)}{s} \underset{\mathrm{(\mathfrak{b})}}{\geqslant} \frac{64\left[\log\left( \frac{64}{\Delta^2}\right) + 2\log\log\left( \frac{64}{\Delta^2} \right) \right]}{s} = \frac{64\log s}{s},
\end{align*}
where $(\mathfrak{b})$ follows since the function $g(x) := x^2 - x - 2\log x$ is monotone increasing for $x \geqslant \log 64$ (think of $\log\left( {64}/{\Delta^2}\right)$ as $x$), and therefore attains its minimum at $x = \log 64$; one can verify that this minimum is strictly positive. Furthermore, since $\log s/s$ is monotone decreasing for $s \geqslant 64$, it follows that for any $m \geqslant T_0$, 
\begin{align*}
\Delta^2 \geqslant \frac{64\log m}{m}.
\end{align*} \hfill $\square$

Now coming back to the proof of Proposition~\ref{prop}, consider an arbitrary $l\in\mathbb{N}$ such that $l > T_0$. Then,
\begin{align*}
\mathbb{P}\left(\tau \leqslant l \right) &= \mathbb{P}\left( \tau \leqslant l, \mathcal{Z} > f\left(T_0\right) \right) + \mathbb{P}\left( \tau \leqslant l, \mathcal{Z} \leqslant f\left(T_0\right) \right) \\
&\leqslant \mathbb{P}\left( \tau \leqslant l, \mathcal{Z} > f\left(T_0\right) \right) + \Phi\left( f\left(T_0\right) \right).
\end{align*}

Now,
\begin{align*}
\mathbb{P}\left( \tau \leqslant l, \mathcal{Z} > f\left(T_0\right) \right) &= \mathbb{P}\left( \bigcup_{m=1}^{l}\left\lbrace \mathcal{Z} + \sum_{j=1}^{m}\left( X_{1,j} - X_{2,j} \right) < 4\sqrt{m\log m}, \mathcal{Z} > f\left(T_0\right) \right\rbrace \right) \\
&\underset{\mathrm{(\dag)}}{=} \mathbb{P}\left( \bigcup_{m = T_0}^{l}\left\lbrace \mathcal{Z} + \sum_{j=1}^{m}\left( X_{1,j} - X_{2,j} \right) < 4\sqrt{m\log m}, \mathcal{Z} > f\left(T_0\right) \right\rbrace \right) \\
&\leqslant \sum_{m=T_0}^{l} \mathbb{P}\left( \mathcal{Z} + \sum_{j=1}^{m}\left( X_{1,j} - X_{2,j} \right) < 4\sqrt{m\log m}, \mathcal{Z} > f\left(T_0\right) \right) \\
&= \sum_{m=T_0}^{l} \mathbb{P}\left( \mathcal{Z} + \sum_{j=1}^{m}\left( X_{1,j} - X_{2,j} -\Delta \right) < -m\left( \Delta - 4\sqrt{\frac{\log m}{m}} \right), \mathcal{Z} > f\left(T_0\right) \right) \\
&\leqslant \sum_{m=T_0}^{l} \mathbb{P}\left( \sum_{j=1}^{m}\left( X_{1,j} - X_{2,j} -\Delta \right) < -m\left( \Delta - 4\sqrt{\frac{\log m}{m}} \right), \mathcal{Z} > f\left(T_0\right) \right) \\
&\underset{\mathrm{(\ddag)}}{\leqslant} \sum_{m=T_0}^{l} \mathbb{P}\left( \sum_{j=1}^{m}\left( X_{1,j} - X_{2,j} -\Delta \right) < -4\sqrt{m\log m}, \mathcal{Z} > f\left(T_0\right) \right) \\
&= \bar{\Phi}\left( f\left(T_0\right) \right)\sum_{m=T_0}^{l} \mathbb{P}\left( \sum_{j=1}^{m}\left( X_{1,j} - X_{2,j} -\Delta \right) < -4\sqrt{m\log m}\right) \\
&\underset{\mathrm{(\star)}}{\leqslant} \bar{\Phi}\left( f\left(T_0\right) \right)\sum_{m=T_0}^{l}\frac{1}{m^8} \\
&\underset{\mathrm{(\ast)}}{\leqslant} \bar{\Phi}\left( f\left(T_0\right) \right)\sum_{m=2}^{\infty}\frac{1}{m^8} \\
&\underset{\mathrm{(\bullet)}}{=} \bar{\Phi}\left( f\left(T_0\right) \right)\left(\zeta(8)-1\right) \\
&\leqslant \frac{\bar{\Phi}\left( f\left(T_0\right) \right)}{200},
\end{align*}
where $(\dag)$ follows from Lemma~\ref{lemma:prop1}, $(\ddag)$ from Lemma~\ref{lemma:prop2}, $(\star)$ follows using the Chernoff-Hoeffding bound \cite{hoeffding}, $(\ast)$ since $T_0 > 1$ (this is because $\Delta \leqslant 1$) and finally in $(\bullet)$, $\zeta(\cdot)$ represents the Riemann zeta function.
Therefore, we have
\begin{align*}
&\mathbb{P}\left(\tau \leqslant l \right) \leqslant \frac{\bar{\Phi}\left( f\left(T_0\right) \right)}{200} + \Phi\left( f\left(T_0\right) \right) = 1 - \left(\frac{199}{200}\right)\bar{\Phi}\left( f\left(T_0\right) \right) \\
\implies\ & \mathbb{P}\left(\tau > l \right) \geqslant \left(\frac{199}{200}\right)\bar{\Phi}\left( f\left(T_0\right) \right).
\end{align*}
Taking the limit $l\to\infty$ and appealing to the continuity of probability, we obtain
\begin{align*}
&\mathbb{P}\left(\tau =\infty \right) \geqslant \left(\frac{199}{200}\right)\bar{\Phi}\left( f\left(T_0\right) \right) \\
\implies\ &\mathbb{P}\left( \bigcap_{m \geqslant 1}\left\lbrace \left\lvert \mathcal{Z} + \sum_{j=1}^{m}\left( X_{1,j} - X_{2,j} \right) \right\rvert \geqslant 4\sqrt{m\log m} \right\rbrace \right) \geqslant \left(\frac{199}{200}\right)\bar{\Phi}\left( f\left(T_0\right) \right).
\end{align*} \hfill $\square$

\section{Analysis of Algorithm~\ref{alg:ALG1}}
\label{sec:analysis-alg1}

The horizon of play is divided into epochs of length $2m = 2\left\lceil \left(2/\delta^2\right) \log n \right\rceil$ each (exactly one pair of arms is played $m$ times each per epoch), e.g., epoch $1$ starts at $t=1$, epoch $2$ at $t=2m+1$, and so on. The decision to \emph{commit} forever to an empirically superior arm or to discard the consideration set of arms and reinitialize the policy, is taken after an epoch ends. For each $k\geqslant 1$, let $S_k$ denote the cumulative pseudo-regret incurred by Algorithm~1 when it is initialized at the beginning of epoch $k$ and run until the end of the horizon, i.e., from $t=(2k-2)m+1$ to $t=n$. Let $S_k^{\prime}$ denote an i.i.d. copy of $S_k$. We are interested in an upper bound on $\mathbb{E}R_n^{\pi} = \mathbb{E}S_1$, where $\pi=$ Algorithm~1. To this end, suppose that $\pi$ is initialized at time $(2k-2)m+1$ (beginning of epoch $k$). Label the arms played in this epoch as $\{1,2\}$ (arm~$1$ is played first). For $i\in\{1,2\}$, let $\bar{X}_i$ denote the empirical mean reward from the $m$ plays of arm $i$ in this epoch. Let $\kappa(i)\in\mathcal{K}=\{1,2\}$ denote the type of arm $i$. Recall that type~$1$ is optimal and that, the probability of a \emph{new} arm queried from the reservoir at time $t$ being optimal-typed is $\alpha(t)$. Let $\mathbbm{1}\{E\}$ denote the indicator corresponding to an event $E$. Consider the following events:
\begin{align}
A &:= \left\lbrace \kappa(1)=1, \kappa(2)=2 \right\rbrace, \\
B &:= \left\lbrace \kappa(1)=2, \kappa(2)=1 \right\rbrace, \\
C &:= \left\lbrace \kappa(1)=2, \kappa(2)=2 \right\rbrace, \\
D &:= \left\lbrace \kappa(1)=1, \kappa(2)=1 \right\rbrace,
\end{align}
where $\kappa(1)$ and $\kappa(2)$ are independent random variables with distributions given by $\mathbb{P}\left( \kappa(1)=1 \right) = \alpha\left( \left(2k-2\right)m+1 \right) =: \alpha_k$ and $\mathbb{P}\left( \kappa(2)=1 \right) = \alpha\left( \left(2k-2\right)m+2 \right) =: \tilde{\alpha}_k$ respectively. Now, observe that $S_k$ evolves according to the following stochastic recursion:
\begin{align}
S_k =\ &\mathbbm{1}\{A\} \left[ \Delta m + \mathbbm{1}\{ \bar{X}_2 - \bar{X}_1 > \delta\} \Delta\left(n-2km\right) + \{ |\bar{X}_1 - \bar{X}_2| < \delta\} S_{k+1}^{\prime} \right] \notag \\
&+\mathbbm{1}\{B\} \left[ \Delta m + \mathbbm{1}\{ \bar{X}_1 - \bar{X}_2 > \delta\} \Delta\left(n-2km\right) + \mathbbm{1}\{ |\bar{X}_1 - \bar{X}_2| < \delta\} S_{k+1}^{\prime} \right] \notag \\
&+\mathbbm{1}\{C\}\left[ 2\Delta m + \mathbbm{1}\{ |\bar{X}_1 - \bar{X}_2| > \delta\}\Delta\left(n-2km\right) + \mathbbm{1}\{ |\bar{X}_1 - \bar{X}_2| < \delta\}S_{k+1}^{\prime} \right] \notag \\
&+\mathbbm{1}\{D\}\mathbbm{1}\{ |\bar{X}_1 - \bar{X}_2| < \delta\}S_{k+1}^{\prime}. \label{eqn:arbityoyo}
\end{align}
Collecting like terms in \eqref{eqn:arbityoyo} together, 
\begin{align}
S_k =\ &\mathbbm{1}\{A\}\mathbbm{1}\{ \bar{X}_2 - \bar{X}_1 > \delta\} \Delta\left( n-2km \right) +\mathbbm{1}\{B\}\mathbbm{1}\{ \bar{X}_1 - \bar{X}_2 > \delta\} \Delta\left( n-2km \right) \notag\\
&+\mathbbm{1}\{C\}\mathbbm{1}\{ |\bar{X}_1 - \bar{X}_2| > \delta\} \Delta\left( n-2km \right) +\left[ \mathbbm{1}\{A \cup B\} + 2\mathbbm{1}\{C\} \right] \Delta m +\mathbbm{1}\{ |\bar{X}_1 - \bar{X}_2| < \delta\}S_{k+1}^{\prime}. \label{eqn:take_exp}
\end{align}
Define the following conditional events:
\begin{align}
E_1 &:= \left\lbrace \left.\bar{X}_2 - \bar{X}_1 > \delta\ \right\rvert\ A \right\rbrace, \label{eqn:e1} \\
E_2 &:= \left\lbrace \left.\bar{X}_1 - \bar{X}_2 > \delta\ \right\rvert\ B \right\rbrace, \label{eqn:e2} \\
E_3 &:= \left\lbrace \left.\left\lvert \bar{X}_1 - \bar{X}_2 \right\rvert > \delta\ \right\rvert\ C \right\rbrace, \label{eqn:e3} \\
E_4 &:= \left\lbrace \left.\left\lvert \bar{X}_1 - \bar{X}_2 \right\rvert < \delta\ \right\rvert\ C \cup D \right\rbrace, \label{eqn:e4} \\
E_5 &:= \left\lbrace \left.\left\lvert \bar{X}_1 - \bar{X}_2 \right\rvert < \delta\ \right\rvert\ A \cup B \right\rbrace. \label{eqn:e5}
\end{align}
Taking expectations on both sides in \eqref{eqn:take_exp} and rearranging, one obtains the following using \eqref{eqn:e1},\eqref{eqn:e2},\eqref{eqn:e3},\eqref{eqn:e4},\eqref{eqn:e5}:
\begin{align}
\mathbb{E}S_k &= \left[ \alpha_k\left(1-\tilde{\alpha}_k\right) \mathbb{P}(E_1) + \tilde{\alpha}_k\left(1-\alpha_k\right) \mathbb{P}(E_2) \right]\Delta\left( n-2km \right) + \left[ \left(1-\alpha_k\right)\left(1-\tilde{\alpha}_k\right)\mathbb{P}(E_3) \right] \Delta\left( n-2km \right) \notag \\
&{\color{white}=} +\left[ \alpha_k\left(1-\tilde{\alpha}_k\right) + \tilde{\alpha}_k\left(1-\alpha_k\right) + 2\left(1-\alpha_k\right)\left(1-\tilde{\alpha}_k\right) \right] \Delta m + \mathbb{P}\left( \left\lvert \bar{X}_1 - \bar{X}_2 \right\rvert < \delta \right) \mathbb{E}S_{k+1}. \label{eqn:take_exp1}
\end{align}
Note that \eqref{eqn:take_exp1} follows from \eqref{eqn:take_exp} since $\mathbb{E}\left[ \mathbbm{1}\{ |\bar{X}_1 - \bar{X}_2| < \delta\}S_{k+1}^{\prime} \right] = \mathbb{P}\left( \left\lvert \bar{X}_1 - \bar{X}_2 \right\rvert < \delta \right) \mathbb{E}S_{k+1}$ due to the independence of $S_{k+1}^{\prime}$. Further note that
\begin{align}
\mathbb{P}\left( \left\lvert \bar{X}_1 - \bar{X}_2 \right\rvert < \delta \right) =\ &\left[ \alpha_k\tilde{\alpha}_k + \left(1-\alpha_k\right)\left(1-\tilde{\alpha}_k\right) \right] \mathbb{P}(E_4) + \left[\alpha_k\left(1-\tilde{\alpha}_k\right) + \tilde{\alpha}_k\left(1-\alpha_k\right)\right]\mathbb{P}(E_5). \label{eqn:take_exp2}
\end{align}
From \eqref{eqn:take_exp1} and \eqref{eqn:take_exp2}, it follows after a little rearrangement that
\begin{align}
\mathbb{E}S_k = \xi_1(k) - k\xi_2(k) + \xi_3(k)\mathbb{E}S_{k+1}, \label{eqn:agp}
\end{align}
where the $\xi_i(k)$'s are given by
\begin{align}
\xi_1(k) &:= \Delta \left[ \alpha_k\left(1-\tilde{\alpha}_k\right) \mathbb{P}(E_1) + \tilde{\alpha}_k\left(1-\alpha_k\right) \mathbb{P}(E_2) \right] n + \Delta \left[\left(1-\alpha_k\right)\left(1-\tilde{\alpha}_k\right)\mathbb{P}(E_3) \right] n + \Delta\left( 2 - \alpha_k -\tilde{\alpha}_k \right) m, \label{eqn:xi1} \\
\xi_2(k) &:= 2\Delta\left[ \alpha_k\left(1-\tilde{\alpha}_k\right) \mathbb{P}(E_1) + \tilde{\alpha}_k\left(1-\alpha_k\right) \mathbb{P}(E_2) \right] m + 2\Delta\left[\left(1-\alpha_k\right)\left(1-\tilde{\alpha}_k\right)\mathbb{P}(E_3) \right] m, \label{eqn:xi2} \\
\xi_3(k) &:= \left[ \alpha_k\tilde{\alpha}_k + \left(1-\alpha_k\right)\left(1-\tilde{\alpha}_k\right) \right] \mathbb{P}(E_4) +\left[\alpha_k\left(1-\tilde{\alpha}_k\right) + \tilde{\alpha}_k\left(1-\alpha_k\right)\right]\mathbb{P}(E_5). \label{eqn:xi3}
\end{align}
Observe that the recursion in \eqref{eqn:agp} is solvable in closed-form and admits the following solution:
\begin{align}
\mathbb{E}S_1 = &\sum_{k=1}^{l}\left(\xi_1(k)\prod_{j=1}^{k-1}\xi_3(j)\right) - \sum_{k=1}^{l}\left(k\xi_2(k)\prod_{j=1}^{k-1}\xi_3(j)\right) + \mathbb{E}S_{l+1}\left(\prod_{k=1}^{l}\xi_3(k)\right), \label{eqn:agp1}
\end{align}
where $l := \left\lfloor n/(2m) \right\rfloor$, $\left\lfloor \cdot \right\rfloor$ being the ``floor'' operator. Since the $\xi_i(k)$'s are non-negative for all $i\in\{1,2,3\}$, $k\in\mathbb{N}$ and $\mathbb{E}S_{l+1} \leqslant 2\Delta m$, it follows that
\begin{align}
\mathbb{E}R_n^{\pi} = \mathbb{E}S_1 \leqslant \sum_{k=1}^{l}\left(\xi_1(k)\prod_{j=1}^{k-1}\xi_3(j)\right) + 2\Delta m, \label{eqn:final_UB}
\end{align}
where the inequality follows since $\xi_3(k)$ is a convex combination of $\mathbb{P}(E_4)$ and $\mathbb{P}(E_5)$ (see \eqref{eqn:xi3}); hence $\xi_3(k) \leqslant 1\ \forall\ k\in\mathbb{N}$. Now using \eqref{eqn:e1},\eqref{eqn:e2},\eqref{eqn:e3},\eqref{eqn:e4},\eqref{eqn:e5} and Hoeffding's inequality \cite{hoeffding} together with the fact that the rewards are bounded in $[0,1]$, we conclude
\begin{align}
\left\lbrace \mathbb{P}(E_1), \mathbb{P}(E_2) \right\rbrace \leqslant \exp\left( -(\Delta+\delta)^2m/2 \right), \label{eqn:hoeff1} \\
\left\lbrace \mathbb{P}(E_3), \mathbb{P}(E_4^c) \right\rbrace \leqslant 2\exp\left( -\delta^2m/2 \right), \label{eqn:hoeff2} \\
\mathbb{P}(E_5) \leqslant \exp\left( -(\Delta-\delta)^2m/2 \right). \label{eqn:hoeff3}
\end{align}
From \eqref{eqn:xi1}, \eqref{eqn:hoeff1} and \eqref{eqn:hoeff2}, it follows that
\begin{align}
\xi_1(k) \leqslant 2\Delta\exp\left( -\delta^2m/2 \right)n + 2\Delta m \leqslant 2\Delta + 2\Delta m, \label{eqn:xi1_upper_bound}
\end{align}
where the last inequality follows since $m = \left\lceil \left(2/\delta^2\right) \log n \right\rceil$, $\left\lceil \cdot \right\rceil$ being the ``ceiling'' operator. Using \eqref{eqn:final_UB} and \eqref{eqn:xi1_upper_bound}, we now have
\begin{align}
\mathbb{E}R_n^{\pi} \leqslant 2\Delta \left[ 1 + \sum_{k=1}^{l}\prod_{j=1}^{k-1}\xi_3(j) \right] (m+1). \label{eqn:yoyba}
\end{align}
From \eqref{eqn:xi3}, observe that 
\begin{align}
\xi_3(k) \leqslant 1 - \left( \alpha_k + \tilde{\alpha}_k - 2\alpha_k\tilde{\alpha}_k \right)\mathbb{P}\left( E_5^c \right) \leqslant \exp\left[ - \left( \alpha_k + \tilde{\alpha}_k - 2\alpha_k\tilde{\alpha}_k \right)\mathbb{P}\left( E_5^c \right) \right]\ \ \ \ \forall\ k\in\mathbb{N}, \label{eqn:yoyu}
\end{align}
where the last inequality follows since $1-x \leqslant \exp(-x)$. From \eqref{eqn:yoyba} and \eqref{eqn:yoyu}, we obtain
\begin{align*}
\mathbb{E}R_n^{\pi} &\leqslant 2\Delta \left[ 1 + \sum_{k=1}^{l}\exp\left( -\mathbb{P}\left( E_5^c \right)\sum_{j=1}^{k-1} \left( \alpha_j + \tilde{\alpha}_j - 2\alpha_j\tilde{\alpha}_j \right) \right) \right] (m+1).
\end{align*}
Recall from \eqref{eqn:hoeff3} that $\mathbb{P}\left( E_5^c \right) \geqslant 1 - \exp\left( -(\Delta-\delta)^2m/2 \right)$. Since $m = \left\lceil \left(2/\delta^2\right) \log n \right\rceil$, it follows that $\exp\left( -(\Delta-\delta)^2m/2 \right) < 1/2$ for $n > 2^{\left(\frac{\delta}{\Delta-\delta}\right)^2}$. Therefore, for $n$ large enough, we have
\begin{align}
\mathbb{E}R_n^{\pi} &\leqslant 2\Delta \left[ 1 + \sum_{k=1}^{l}\exp\left( -\frac{1}{2}\sum_{j=1}^{k-1} \left( \alpha_j + \tilde{\alpha}_j - 2\alpha_j\tilde{\alpha}_j \right) \right) \right] (m+1) \notag \\
&\leqslant 2\Delta \left[ 3 + \sum_{k=3}^{l}\exp\left( -\frac{1}{2}\sum_{j=1}^{k-1} \left( \alpha_j + \tilde{\alpha}_j - 2\alpha_j\tilde{\alpha}_j \right) \right) \right] (m+1) \notag \\ 
&\leqslant 2\Delta \left[ 3 + \sum_{k=3}^{l}\exp\left( -\frac{1}{2}\sum_{j=2}^{k-1} \left( \alpha_j + \tilde{\alpha}_j - 2\alpha_j\tilde{\alpha}_j \right) \right) \right] (m+1). \label{eqn:yoyoyuyu}
\end{align}

This concludes the basic analysis of Algorithm~1. We will use these results in subsequent sub-sections to provide the proofs for specific functional forms of $\alpha(t)$.

\subsection{Proof of Theorem~\ref{thm:alg1-exo}}
\label{proof:alg1-exo}

Recall that $\alpha_j := \alpha\left( (2j-2)m+1\right)$. Since $\alpha(t) \sim ct^{-\gamma}$, it follows that for $n$ large enough (equivalently, $m$ large enough since $m = \left\lceil \left(2/\delta^2\right) \log n \right\rceil$), we have $\alpha_j \leqslant 1/2$ for all $j \geqslant 2$, which implies $\alpha_j + \tilde{\alpha}_j - 2\alpha_j\tilde{\alpha}_j \geqslant \alpha_j$ for all $j \geqslant 2$. Therefore, it follows from \eqref{eqn:yoyoyuyu} that for $n$ large enough,
\begin{align}
\mathbb{E}R_n^\pi \leqslant 2\Delta \left[ 3 + \sum_{k=3}^{l}\exp\left( -\frac{1}{2}\sum_{j=2}^{k-1} \alpha_j \right) \right] (m+1). \label{eqn:yu-temp}
\end{align}
Now, since $\alpha(t) \sim ct^{-\gamma}$, it follows that for $n$ large enough (equivalently, $m$ large enough),
\begin{align}
\alpha_j \geqslant \frac{c}{(2jm)^\gamma}, \label{eqn:alphas}
\end{align}
Combining \eqref{eqn:yu-temp} and \eqref{eqn:alphas}, we get that for $n$ large enough,
\begin{align}
\mathbb{E}R_n^{\pi} &\leqslant 2\Delta \left[ 3 + \sum_{k=3}^{l}\exp\left( -\frac{c}{m^{\gamma}2^{\gamma+1}}\sum_{j=2}^{k-1}j^{-\gamma} \right) \right] (m+1) \label{eqn:yelo} \\
&\leqslant 2\Delta \left[ 3 + \sum_{k=3}^{l}\exp\left( -\frac{c}{m^{\gamma}2^{\gamma+1}}\int_{2}^{k}x^{-\gamma}dx \right) \right] (m+1) \notag \\
&= 2\Delta \left[ 3 + \sum_{k=3}^{l}\exp\left( -\frac{c\left( k^{1-\gamma} - 2^{1-\gamma} \right) }{(1-\gamma)m^{\gamma}2^{\gamma+1}} \right) \right] (m+1) \notag \\
&\leqslant 2\Delta \left[ 3 + \sum_{k=3}^{l}\exp\left( -\frac{c\left( k^{1-\gamma} - 2^{1-\gamma} \right) }{4(1-\gamma)m^{\gamma}} \right) \right] (m+1) \notag \\
&\leqslant 6\Delta \left[ 1 + \sum_{k=3}^{l}\exp\left( -\frac{c k^{1-\gamma} }{4(1-\gamma)m^{\gamma}} \right) \right] (m+1), \label{eqn:len}
\end{align}
where the last inequality holds since $m = \left\lceil \left(2/\delta^2\right) \log n \right\rceil$ and $n$ is large enough. Now observe that
\begin{align}
\mathbb{E}R_n^{\pi} &\leqslant 6\Delta \left[ 1 + \sum_{k=3}^{l}\exp\left( -\frac{c k^{1-\gamma} }{4(1-\gamma)m^{\gamma}} \right) \right] (m+1) \notag \\
&\leqslant 6\Delta \left[ 1 + \int_{2}^{l}\exp\left( -\frac{c x^{1-\gamma} }{4(1-\gamma)m^{\gamma}} \right)dx \right] (m+1) \notag \\
&\leqslant 6\Delta \left[ 1 + \int_{2}^{n/(2m)}\exp\left( -\frac{c x^{1-\gamma} }{4(1-\gamma)m^{\gamma}} \right)dx \right] (m+1), \label{eqn:finalyoyu}
\end{align}
where the last inequality follows since $l = \left\lfloor n/(2m) \right\rfloor$. We now focus on solving the integral. Define
\begin{align*}
\mathcal{I} &:= \int_{2}^{n/(2m)}\exp\left( -\frac{c x^{1-\gamma} }{4(1-\gamma)m^{\gamma}} \right)dx \\
&\leqslant \int_{0}^{\infty}\exp\left( -\frac{c x^{1-\gamma}  }{4(1-\gamma)m^{\gamma}} \right)dx \\
&\underset{\mathrm{(\ddag)}}{=} (1-\gamma)^{\frac{\gamma}{1-\gamma}}\left( \frac{4m^\gamma}{c} \right)^{\frac{1}{1-\gamma}}\int_{0}^{\infty}z^{\frac{\gamma}{1-\gamma}}\exp(-z)dz \\
&\leqslant \left( \frac{4}{c} \right)^{\frac{1}{1-\gamma}}\left( \int_{0}^{\infty}z^{\frac{\gamma}{1-\gamma}}\exp(-z)dz \right) m^{\frac{\gamma}{1-\gamma}} \\
&\leqslant \left( \frac{4}{c} \right)^{\frac{1}{1-\gamma}}\left( \int_{0}^{\infty}z^{\left\lceil \frac{\gamma}{1-\gamma} \right\rceil}\exp(-z)dz \right) m^{\frac{\gamma}{1-\gamma}},
\end{align*}
where $(\ddag)$ follows after the variable substitution $z = \frac{cx^{1-\gamma}}{4(1-\gamma)m^{\gamma}}$. Now, the $\left\lceil \frac{\gamma}{1-\gamma} \right\rceil^{\text{th}}$ moment of a unit rate exponential random variable is given by the \emph{factorial} of $\left\lceil \frac{\gamma}{1-\gamma} \right\rceil$, denoted by $\mathfrak{F}\left(\left\lceil \frac{\gamma}{1-\gamma} \right\rceil\right)$. Thus, we have
\begin{align}
\mathcal{I} \leqslant \left( \frac{4}{c} \right)^{\frac{1}{1-\gamma}}\mathfrak{F}\left(\left\lceil \frac{\gamma}{1-\gamma} \right\rceil\right) m^{\frac{\gamma}{1-\gamma}}. \label{eqn:finalfinal}
\end{align}
Combining \eqref{eqn:finalyoyu} and \eqref{eqn:finalfinal}, we conclude that for large enough $n$,
\begin{align*}
\mathbb{E}R_n^{\pi} \leqslant 24\Delta \left( \frac{4}{c} \right)^{\frac{1}{1-\gamma}}\mathfrak{F}\left(\left\lceil \frac{\gamma}{1-\gamma} \right\rceil\right) m^{\frac{1}{1-\gamma}}.
\end{align*}
Finally, since $m = \left\lceil \left(2/\delta^2\right) \log n \right\rceil$, the stated assertion follows, i.e.,
\begin{align*}
\limsup_{n\to\infty}\frac{\mathbb{E}R_n^{\pi}}{\left(\log n\right)^{\frac{1}{1-\gamma}}} \leqslant 24\Delta\left( \frac{8}{\delta^2c} \right)^{\frac{1}{1-\gamma}} \mathfrak{F}\left(\left\lceil \frac{\gamma}{1-\gamma} \right\rceil\right).
\end{align*} \hfill $\square$

\subsection{Proof of Theorem~\ref{thm:alg1-endo}}
\label{proof:alg1-endo}

Again, we pick things up from \eqref{eqn:yoyoyuyu}. We know that
\begin{align*}
\mathbb{E}R_n^{\pi} \leqslant 2\Delta \left[ 3 + \sum_{k=3}^{l}\exp\left( -\frac{1}{2}\sum_{j=2}^{k-1} \left( \alpha_j + \tilde{\alpha}_j - 2\alpha_j\tilde{\alpha}_j \right) \right) \right] (m+1),
\end{align*}
where $m = \left\lceil \left(2/\delta^2\right) \log n \right\rceil$, $\alpha_j = \alpha\left(\left(2j-2\right)m+1\right)$ and $\tilde{\alpha}_j = \alpha\left(\left(2j-2\right)m+2\right)$. Since $\alpha(t) = g\left( \mathcal{J}_{t-1}\right)$, we have $\alpha_j = g\left( 2(j-1) \right)$ and $\tilde{\alpha}_j = g\left( 2j-1 \right)$. Since $g(\cdot) \leqslant c \leqslant 1/2$, it follows that $\alpha_j + \tilde{\alpha}_j - 2\alpha_j\tilde{\alpha}_j \geqslant \alpha_j = g\left( 2(j-1) \right)$ for all $j \geqslant 1$. Therefore, one has 
\begin{align}
\mathbb{E}R_n^\pi &\leqslant 2\Delta \left[ 3 + \sum_{k=3}^{l}\exp\left( -\frac{1}{2}\sum_{j=2}^{k-1} g\left( 2(j-1) \right) \right) \right] (m+1) \notag \\
&= 2\Delta \left[ 3 + \sum_{k=3}^{l}\exp\left( -\frac{c}{2}\sum_{j=2}^{k-1} \left( 2j-1 \right)^{-\gamma} \right) \right] (m+1) \notag \\
&\leqslant 2\Delta \left[ 3 + \sum_{k=3}^{l}\exp\left( -\frac{c}{2^{\gamma+1}}\sum_{j=2}^{k-1} j^{-\gamma} \right) \right] (m+1). \label{eqn:idhar_wala}
\end{align}

Drawing upon structural similarities between \eqref{eqn:idhar_wala} and \eqref{eqn:yelo}, one can proceed along an analogous sequence of steps to eventually conclude
\begin{align*}
\limsup_{n\to\infty}\frac{\mathbb{E}R_n^{\pi}}{\log n} \leqslant \left( \frac{48\Delta}{\delta^2} \right) \left( \frac{4}{c} \right)^{\frac{1}{1-\gamma}} \mathfrak{F}\left(\left\lceil \frac{\gamma}{1-\gamma} \right\rceil\right),
\end{align*} 
where $\mathfrak{F}(\cdot)$ denotes the \emph{Factorial} function. \hfill $\square$

\section{Auxiliary results used in the analysis of Algorithm~\ref{alg:ALG2}}
\label{sec:aux-alg2}


\begin{fact}[Lemma~2 of \cite{kalvit2020finite}]
\label{fact:2}

Consider a stochastic two-armed bandit with $[0,1]$-valued rewards and gap $\Delta$. Let $\left( X_{i,j} \right)_{j \geqslant 1}$ be the reward sequence associated with arm~$i$ and $N_i(n)$ its sample count until time $n$ under UCB1 \cite{auer2002}. Define $M_n := \min\left( N_1(n), N_2(n) \right)$ and consider the following stopping time:
\begin{align*}
T := \inf \left\lbrace n \geqslant 2 : \left\lvert \sum_{j=1}^{M_n} \left( X_{1,j} - X_{2,j} \right) \right\rvert < 4\sqrt{M_n\log M_n} \right\rbrace.
\end{align*}
Then, the following results hold:
\begin{enumerate}
\item If $\Delta>0$, then $\mathbb{P}\left( T=\infty \right) \geqslant \beta_\Delta > 0$, where $\beta_\Delta$ is as defined in \eqref{eqn:beta}.
\item If $\Delta=0$, then $\mathbb{E}T \leqslant C_0 < \infty$, where $C_0$ is some absolute constant.
\end{enumerate}
\end{fact}

\section{Analysis of Algorithm~\ref{alg:ALG2}}
\label{sec:analysis-alg2}

Algorithm~\ref{alg:ALG2} runs in epochs of stochastic durations that are determined online in an adaptive manner. Let the sequence of epoch durations be $\left( T_k : k = 1,2,... \right)$. Define $W_n := \inf \left\lbrace l\in\mathbb{N} : \sum_{k=1}^{l}T_k \geqslant n \right\rbrace$. Let $I_k(i)$ denote the event that each of the two arms queried at the beginning of epoch $k$ has type $i$ ($i$ is an element of $\mathcal{K}=\{1,2\}$). Similarly, let $D_k$ denote the event that the aforementioned arms have ``distinct'' types. Suppose that $S_n$ denotes the cumulative pseudo-regret of UCB1 \cite{auer2002} after $n$ plays in an independent instance of a two-armed bandit with gap $\Delta$. Then, the evolution of the cumulative pseudo-regret of the policy $\pi$ given by Algorithm~\ref{alg:ALG2} after any number $n$ of plays satisfies
\begin{align*}
\tilde{R}_n^{\pi} &= \sum_{k=1}^{W_n-1} \left[ \mathbbm{1}\left\lbrace D_k \right\rbrace S_{T_k} +\mathbbm{1}\left\lbrace I_k(2) \right\rbrace \Delta T_k \right] + \mathbbm{1}\left\lbrace D_{W_n} \right\rbrace S_{\left( n - \sum_{k=1}^{W_n-1}T_k\right)} +\mathbbm{1}\left\lbrace I_{W_n}(2) \right\rbrace \Delta \left( n - \sum_{k=1}^{W_n-1}T_k\right) \\
&\underset{\mathrm{(\dag)}}{\leqslant} \sum_{k=1}^{W_n-1} \left[ \mathbbm{1}\left\lbrace D_k \right\rbrace S_n +\mathbbm{1}\left\lbrace I_k(2) \right\rbrace \Delta T_k \right] + \mathbbm{1}\left\lbrace D_{W_n} \right\rbrace S_n +\mathbbm{1}\left\lbrace I_{W_n}(2) \right\rbrace \Delta T_{W_n} \\
&= \sum_{k=1}^{W_n} \left[ \mathbbm{1}\left\lbrace D_k \right\rbrace S_n +\mathbbm{1}\left\lbrace I_k(2) \right\rbrace \Delta T_k \right] \\
&= \sum_{k=1}^{\infty} \mathbbm{1}\left\lbrace W_n \geqslant k\right\rbrace\mathbbm{1}\left\lbrace D_k \right\rbrace S_n + \sum_{k=1}^{\infty} \mathbbm{1}\left\lbrace W_n \geqslant k\right\rbrace\mathbbm{1}\left\lbrace I_k(2) \right\rbrace \Delta T_k,
\end{align*}
where $(\dag)$ follows since the pseudo-regret of UCB1 is weakly increasing in the horizon, and $n - \sum_{k=1}^{W_n-1}T_k \leqslant T_{W_n}$. Taking expectations and invoking Tonelli's theorem to interchange expectation and infinite-sum (since all summands are non-negative), we obtain
\begin{align}
\mathbb{E}R_n^{\pi} = \mathbb{E}\tilde{R}_n^{\pi} \leqslant \mathbb{E}S_n\sum_{k=1}^{\infty} \mathbb{P}\left( D_k, W_n \geqslant k \right) + \Delta \sum_{k=1}^{\infty}\mathbb{P}\left( I_k(2), W_n \geqslant k \right)\mathbb{E}\left[ T_k\ |\ I_k(2),\ W_n \geqslant k \right]. \label{eqn:regret-raw}
\end{align}

(Note that $\mathbb{E}\left[ \left. S_n \right\rvert W_n \geqslant k,\ D_k \right] = \mathbb{E}S_n$ since $S_n$, by definition, is independent of $W_n$ and $D_k$.)

\begin{lemma}[Bounded epochs for identical-typed arms]
\label{lemma:intermediate}
The following holds for any $k,n\in\mathbb{N}$ and arm-type $i\in\{1,2\}$:
\begin{align*}
\mathbb{E}\left[ T_k\ |\ I_k(i),\ W_n \geqslant k \right] = \mathbb{E}\left[ T_k\ |\ I_k(i)\right] \leqslant C_0 < \infty,
\end{align*}
where $C_0$ is as given in Fact~\ref{fact:2}.
\end{lemma}

\textit{Proof of Lemma~\ref{lemma:intermediate}.} Note that
\begin{align*}
&\mathbb{E}\left[ T_k\ |\ I_k(i) \right] \\
=\ &\mathbb{E}\left[ T_k\ |\ I_k(i),\ W_n \geqslant k \right]\mathbb{P}\left( W_n \geqslant k\ |\ I_k(i) \right) + \mathbb{E}\left[ T_k\ |\ I_k(i),\ W_n < k \right]\mathbb{P}\left( W_n < k\ |\ I_k(i) \right) \\
\underset{\mathrm{(\dag)}}{=}\ &\mathbb{E}\left[ T_k\ |\ I_k(i),\ W_n \geqslant k \right]\mathbb{P}\left( W_n \geqslant k\ |\ I_k(i) \right) + \mathbb{E}\left[ T_k\ |\ I_k(i)\right]\mathbb{P}\left( W_n < k\ |\ I_k(i) \right),
\end{align*}
where $(\dag)$ follows since $T_k$ is independent of $W_n$, given $I_k(i)$ and $k>W_n$. Thus,
\begin{align*}
\mathbb{E}\left[ T_k\ |\ I_k(i),\ W_n \geqslant k \right]= \mathbb{E}\left[ T_k\ |\ I_k(i)\right] \leqslant C_0,
\end{align*}
with the last inequality following from Fact~\ref{fact:2}.2. \hfill $\square$

Now coming back to the analysis of Algorithm~\ref{alg:ALG2}, observe that from \eqref{eqn:regret-raw} and Lemma~\ref{lemma:intermediate}, one has
\begin{align}
\mathbb{E}R_n^{\pi} \leqslant \mathbb{E}S_n\sum_{k=1}^{\infty} \mathbb{P}\left( D_k, W_n \geqslant k \right) + C_0\Delta \sum_{k=1}^{\infty}\mathbb{P}\left( I_k(2), W_n \geqslant k \right). \label{eqn:regret-raw1}
\end{align}

Note that
\begin{align}
\mathbb{P}\left( W_n \geqslant k \right) &= \mathbb{P}\left( \sum_{m=1}^{k-1} T_m < n \right) \notag \\
&= \mathbb{P}\left( \bigcap_{j=1}^{k-1} \left\lbrace \sum_{m=1}^{j} T_m < n \right\rbrace \right) \notag \\
&= \mathbb{P}\left( T_1 < n \right)\prod_{j=2}^{k-1}\mathbb{P}\left( \left. \sum_{m=1}^{j}T_m < n\ \right\rvert\ T_1 < n,\ ...,\ \sum_{m=1}^{j-1}T_m < n \right) \notag \\
&= \mathbb{P}\left( T_1 < n \right)\prod_{j=2}^{k-1}\mathbb{P}\left( \left. \sum_{m=1}^{j}T_m < n\ \right\rvert\ \sum_{m=1}^{j-1}T_m < n \right) \notag \\
&\leqslant \mathbb{P}\left( T_1 < \infty \right)\prod_{j=2}^{k-1}\mathbb{P}\left( T_j < \infty \ \left\lvert\ \sum_{m=1}^{j-1}T_m < n \right. \right). \label{eqn:Wynamic}
\end{align}

This concludes the basic analysis of Algorithm~\ref{alg:ALG2}. We will use these results in subsequent sub-sections to prove guarantees under specific functional forms of $\alpha(t)$.

\subsection{Proof of Theorem~\ref{thm:alg2-exo}}
\label{proof:alg2-exo}

In this setting, $\left( \alpha(t) : t = 1,2,... \right)$ is a non-increasing process. Observe that conditional on the event $E_l:=\left\lbrace \sum_{m=1}^{j-1}T_m = l \right\rbrace$, where $l<n$ is arbitrary, the probability that $T_j < \infty$ (for $j\geqslant 2$), satisfies 
\begin{align}
\mathbb{P}\left( T_j < \infty \left\lvert E_l \right. \right) &= \mathbb{P}\left( T_j<\infty\ |\ E_l, I_j(1)\cup I_j(2) \right)\mathbb{P}\left( I_j(1)\cup I_j(2) | E_l \right) + \mathbb{P}\left( T_j<\infty\ |\ E_l, D_j \right)\mathbb{P}\left( D_j | E_l \right) \notag \\
&\underset{\mathrm{(\dag)}}{\leqslant} \mathbb{P}\left( I_j(1)\cup I_j(2) | E_l \right) + \mathbb{P}\left( T_j<\infty\ |\ D_j \right)\mathbb{P}\left( D_j | E_l \right) \notag \\
&\underset{\mathrm{(\ddag)}}{\leqslant} \mathbb{P}\left( I_j(1)\cup I_j(2) | E_l \right) + \left( 1 - \beta_\Delta \right) \mathbb{P}\left( D_j | E_l \right) \notag \\
&= 1 - \mathbb{P}\left( D_j | E_l\right)\beta_\Delta, \notag \\
&= 1 - \left[ \alpha(l+1)\left(1-\alpha(l+2)\right) + \alpha(l+2)\left(1-\alpha(l+1)\right) \right]\beta_\Delta \notag \\
&\underset{\mathrm{(\star)}}{\leqslant} 1 - \alpha(l+1)\beta_\Delta \notag \\
&\underset{\mathrm{(\ast)}}{\leqslant} 1 - \alpha(n)\beta_\Delta, \notag
\end{align}
where $(\dag)$ follows because $T_j$ is independent of $E_l$, given $D_j$, and $(\ddag)$ follows using Fact~\ref{fact:2}.1. Next, $(\star)$ follows since $\alpha(l+1) \leqslant 1/2$ by assumption and finally, $(\ast)$ since $l+1 \leqslant n$ and $\alpha(\cdot)$ is non-increasing. Notice that although $(\ast)$ holds for $j \geqslant 2$, the same upper bound of $1 - \alpha(n)\beta_\Delta$ holds trivially also for $\mathbb{P}\left( T_1 < \infty \right)$ (proof is almost identical to that for $j \geqslant 2$ except that the probabilities are unconditional). Using $(\ast)$ and said observation in \eqref{eqn:Wynamic}, one concludes that
\begin{align}
\mathbb{P}\left( W_n \geqslant k \right) \leqslant \left( 1 - \alpha(n)\beta_\Delta \right)^{k-1}. \label{eqn:W_n-dynamic-yo}
\end{align}

Now, we know from \eqref{eqn:regret-raw1} that 
\begin{align*}
\mathbb{E}R_n^{\pi} \leqslant \mathbb{E}S_n\sum_{k=1}^{\infty}\mathbb{P}\left( W_n \geqslant k \right) + C_0\Delta\sum_{k=1}^{\infty}\mathbb{P}\left( W_n \geqslant k \right) \leqslant \frac{\mathbb{E}S_n + C_0\Delta}{\alpha(n)\beta_\Delta},
\end{align*}
where the last step follows using \eqref{eqn:W_n-dynamic-yo}. Finally, using $\mathbb{E}S_n \leqslant \left( 8/\Delta \right)\log n + \left( 1 + \pi^2/3 \right)\Delta$ \cite{auer2002} and taking appropriate limits, the stated assertion follows. \hfill $\square$

\subsection{Proof of Theorem~\ref{thm:alg2-endo}}
\label{proof:alg2-endo}

In this setting, $\alpha(t) = g\left( \mathcal{J}_{t-1} \right)$, where $\mathcal{J}_{t-1}$ is the number of reservoir queries until time $t-1$ (inclusive) and $g(\cdot)$ is non-increasing. Since the dependence on $t$ is only through $\mathcal{J}_{t-1}$, joint probabilities in \eqref{eqn:regret-raw1} decouple into products and we have the following lemma:

\begin{lemma}[Product-form probabilities]
\label{lemma:intermediate2}
Consider the following cases:
\begin{enumerate}
\item For any $k\in\mathbb{N}$ and any arm-type~$i\in\{1,2\}$, the events $D_k,I_k(i)$ are independent of the ``starting time'' of epoch $k$.
\item For any $k\in\mathbb{N}$ and any arm-type~$i\in\{1,2\}$, the events $D_k,I_k(i)$ depend on the ``starting time'' of epoch $k$ only through $k$.
\end{enumerate}
In either case, one has for any $k,n\in\mathbb{N}$ and any arm-type~$i\in\{1,2\}$ that
\begin{align*}
&\mathbb{P}\left( D_k, W_n \geqslant k \right) = \mathbb{P}\left( D_k \right)\mathbb{P}\left( W_n \geqslant k \right), \\
&\mathbb{P}\left( I_k(i), W_n \geqslant k \right) = \mathbb{P}\left( I_k(i) \right)\mathbb{P}\left( W_n \geqslant k \right).
\end{align*}
\end{lemma}

\textit{Proof of Lemma~\ref{lemma:intermediate2}.} We know that
\begin{align*}
\mathbb{P}\left( D_k, W_n < k \right) &= \mathbb{P}\left( W_n < k \right)\mathbb{P}\left( D_k\ \left\lvert\ k > W_n \right. \right) \\
&= \mathbb{P}\left( W_n < k \right)\mathbb{P}\left( D_k \right) & \mbox{(by assumption)} \\
&= \mathbb{P}\left( D_k \right) - \mathbb{P}\left( D_k \right)\mathbb{P}\left( W_n \geqslant k \right) \\
\implies \mathbb{P}\left( D_k, W_n \geqslant k \right) &= \mathbb{P}\left( D_k \right)\mathbb{P}\left( W_n \geqslant k \right).
\end{align*}
The result for $I_k(i)$ can also be shown similarly. \hfill $\square$

Now coming back to the proof of Theorem~\ref{thm:alg2-endo}, we will pick things up from \eqref{eqn:regret-raw1}. Using Lemma~\ref{lemma:intermediate2} with \eqref{eqn:regret-raw1}, we obtain
\begin{align}
\mathbb{E}R_n^{\pi} &\leqslant \mathbb{E}S_n\sum_{k=1}^{\infty} \mathbb{P}\left( D_k \right) \mathbb{P} \left( W_n \geqslant k \right) + C_0\Delta \sum_{k=1}^{\infty}\mathbb{P}\left( I_k(2) \right) \mathbb{P} \left( W_n \geqslant k \right) \notag \\
&\leqslant \mathbb{E}S_n\sum_{k=1}^{\infty} \left[ g\left(2k-2\right) + g\left( 2k-1 \right) - 2 g\left(2k-2\right) g\left( 2k-1 \right) \right] \mathbb{P} \left( W_n \geqslant k \right) + C_0\Delta \sum_{k=1}^{\infty} \mathbb{P} \left( W_n \geqslant k \right) \notag \\
&\underset{\mathrm{(\$)}}{\leqslant} 2\mathbb{E}S_n\sum_{k=1}^{\infty} g\left(2k-2\right) \mathbb{P} \left( W_n \geqslant k \right) + C_0\Delta \sum_{k=1}^{\infty} \mathbb{P} \left( W_n \geqslant k \right) \notag \\
&= 2\mathbb{E}S_n\sum_{k=0}^{\infty} g\left(2k\right) \mathbb{P} \left( W_n \geqslant k+1 \right) + C_0\Delta \sum_{k=1}^{\infty} \mathbb{P} \left( W_n \geqslant k \right), \label{eqn:ye-le-final-wala}
\end{align}
where $(\$)$ follows since $g(\cdot)$ is non-increasing. Next, we upper bound $\mathbb{P} \left( W_n \geqslant k \right)$. Observe that conditional on the event $E_l:=\left\lbrace \sum_{m=1}^{j-1}T_m = l \right\rbrace$, where $l<n$ is arbitrary, the probability that $T_j < \infty$ (for $j\geqslant 2$), satisfies 
\begin{align}
\mathbb{P}\left( T_j < \infty \left\lvert E_l \right. \right) &= \mathbb{P}\left( T_j<\infty\ |\ E_l, I_j(1)\cup I_j(2) \right)\mathbb{P}\left( I_j(1)\cup I_j(2) | E_l \right) + \mathbb{P}\left( T_j<\infty\ |\ E_l, D_j \right)\mathbb{P}\left( D_j | E_l \right) \notag \\
&\underset{\mathrm{(\dag)}}{\leqslant} \mathbb{P}\left( I_j(1)\cup I_j(2) | E_l \right) + \mathbb{P}\left( T_j<\infty\ |\ D_j \right)\mathbb{P}\left( D_j | E_l \right) \notag \\
&\underset{\mathrm{(\ddag)}}{\leqslant} \mathbb{P}\left( I_j(1)\cup I_j(2) | E_l \right) + \left( 1 - \beta_\Delta \right) \mathbb{P}\left( D_j | E_l \right) \notag \\
&= 1 - \mathbb{P}\left( D_j | E_l\right)\beta_\Delta, \notag \\
&= 1 - \left[ \alpha(l+1)\left(1-\alpha(l+2)\right) + \alpha(l+2)\left(1-\alpha(l+1)\right) \right]\beta_\Delta \notag \\
&\underset{\mathrm{(\star)}}{\leqslant} 1 - \alpha(l+1)\beta_\Delta \notag \\
&\underset{\mathrm{(\ast)}}{=} 1 - g\left(2j-2\right)\beta_\Delta, \notag
\end{align}
where $(\dag)$ follows because $T_j$ is independent of $E_l$, given $D_j$, and $(\ddag)$ follows using Fact~\ref{fact:2}.1. Next, $(\star)$ follows since $\alpha(l+1) \leqslant 1/2$ by assumption and finally, $(\ast)$ holds since $\alpha(l+1) = g\left( \mathcal{J}_l \right) = g\left(2j-2\right)$ on $E_l$. Notice that although $(\ast)$ holds for $j \geqslant 2$, the same upper bound of $1 - g\left(2j-2\right)\beta_\Delta$ holds trivially also for $\mathbb{P}\left( T_1 < \infty \right)$ (proof is almost identical to that for $j \geqslant 2$ except that the probabilities are unconditional). Using $(\ast)$ and said observation in \eqref{eqn:Wynamic}, one concludes that
\begin{align}
\mathbb{P}\left( W_n \geqslant k \right) \leqslant \prod_{j=1}^{k-1}\left( 1 - g\left(2j-2\right)\beta_\Delta \right) = \prod_{j=0}^{k-2}\left( 1 - \beta_\Delta g\left(2j\right) \right). \label{eqn:W_n-dynamic-yo-endo}
\end{align}

Combining \eqref{eqn:ye-le-final-wala} and \eqref{eqn:W_n-dynamic-yo-endo}, one obtains
\begin{align*}
\mathbb{E}R_n^\pi &\leqslant 2\mathbb{E}S_n\sum_{k=0}^{\infty} g\left(2k\right) \prod_{j=0}^{k-1}\left( 1 - \beta_\Delta g\left(2j\right) \right) + C_0\Delta \sum_{k=0}^{\infty} \prod_{j=0}^{k-1}\left( 1 - \beta_\Delta g\left(2j\right) \right) \\
&\underset{\mathrm{(\#_1)}}{\leqslant} \left( 2c\mathbb{E}S_n + C_0\Delta \right)\sum_{k=0}^{\infty} \prod_{j=0}^{k-1}\left( 1 - \beta_\Delta g\left(2j\right) \right) \\
&\underset{\mathrm{(\#_2)}}{\leqslant} \left( 2c\mathbb{E}S_n + C_0\Delta \right)\sum_{k=0}^{\infty} \exp \left( - \beta_\Delta \sum_{j=0}^{k-1} g\left(2j\right) \right),
\end{align*}
where $\left(\#_1\right)$ follows since $g(\cdot)$ is non-increasing with $g(0) = c$, and $\left(\#_2\right)$ follows using the identity $\log(1+x) \leqslant x\ \forall\ x > -1$. Finally, using $\mathbb{E}S_n \leqslant \left( 8/\Delta \right)\log n + \left( 1 + \pi^2/3 \right)\Delta$ \cite{auer2002} and taking appropriate limits, the stated assertion follows. \hfill $\square$

\end{document}